\documentclass[sigconf]{acmart_modified}
\usepackage{booktabs} % For formal tables

\setcopyright{none}
\DeclareMathOperator{\distW}{\mathcal{W}}
\DeclareMathOperator{\calA}{\mathcal{A}}

\newcommand{\onlyinproc}[1]{}
\newcommand{\notinarxive}[1]{}
\newcommand{\onlyinarxive}[1]{#1}
\onlyinproc{
\acmDOI{10.475/123_4}

\acmISBN{123-4567-24-567/08/06}

\acmConference[KDD'17]{ACM}{August 2017}{ USA} 
\acmYear{2017}
\copyrightyear{2017}
\acmPrice{15.00}
}

\usepackage{url}
\usepackage{algorithm}
\usepackage[algo2e, ruled, vlined]{algorithm2e}

\def\boldW{\boldsymbol{W}}
\def\boldD{\boldsymbol{D}}
\def\boldY{\boldsymbol{Y}}
\def\boldI{\boldsymbol{I}}
\def\boldA{\boldsymbol{A}}
\def\boldF{\boldsymbol{F}}
\def\vecf{\boldsymbol{f}}
\def\vecc{\boldsymbol{c}}
\def\vecy{\boldsymbol{y}}
\def\vecw{\boldsymbol{w}}

\def\vec0{\boldsymbol{0}}
\def\Exp{\textsf{Exp}}
\newcommand\E{\textsf{E}}
\newcommand{\ignore}[1]{}

\definecolor{darkgreen}{RGB}{0,100,0}
\newcommand{\edith}[1]{\footnote{\textcolor{purple}{EC}:#1}}

\begin{document}
\title{Bootstrapped Graph Diffusions: \\ Exposing the Power of Nonlinearity}
%\subtitle{KDD '17 submission \#449}

\author{Eliav Buchnik}
% \authornote{Dr.~Trovato insisted his name be first.}
% \orcid{1234-5678-9012}
\affiliation{%
  \institution{Tel Aviv University, Israel}
  \institution{Google Research, Israel}
%  \streetaddress{P.O. Box 1212}
%  \city{Dublin} 
%  \state{Israel} 
%  \postcode{43017-6221}
}
\email{eliavbuh@gmail.com}

\author{Edith Cohen}
% \authornote{Dr.~Trovato insisted his name be first.}
% \orcid{1234-5678-9012}
\affiliation{%
  \institution{Google Research, CA, USA}
%  \state{CA, USA} 
}
\affiliation{
\institution{Tel Aviv University, Israel}
%  \streetaddress{P.O. Box 1212}
%  \city{Dublin} 
%  \state{Israel} 
%  \postcode{43017-6221}
}
\email{edith@cohenwang.com}

%\numberofauthors{2}
% \author{%\alignauthor 
% Eliav Buchnik \\ Tel Aviv Univesrity, Israel
% \email{eliavbuh@gmail.com} \and Edith Cohen\\ 
% Google Research, USA \\ 
%  Tel Aviv University, Israel \\ 
% \email{edith@cohenwang.com}}

 \begin{abstract}
Graph-based semi-supervised learning (SSL) algorithms predict labels
for all nodes based on provided labels of a small set of seed nodes.
Classic methods capture the graph structure through some underlying
diffusion process that propagates through the graph edges.
  Spectral diffusion, which includes personalized page rank
and label propagation, propagates through random walks.  Social
diffusion propagates through shortest paths.
A common ground to these diffusions is their
{\em linearity}, which does not
distinguish between contributions of
few ``strong'' relations and many ``weak'' relations.

Recently,  non-linear methods such as node embeddings  and graph convolutional
networks (GCN) demonstrated a large gain in quality for SSL tasks.
These methods
introduce  multiple components and greatly vary on how the graph
 structure, seed label information, and other features are used.

We aim here to study  the contribution of
non-linearity, as an isolated ingredient, to the performance gain.  To do so, 
we place classic linear graph diffusions in a self-training framework.  
Surprisingly, we observe that SSL using the resulting {\em bootstrapped
diffusions} not only significantly improves over the respective
non-bootstrapped baselines
but also outperform 
state-of-the-art non-linear SSL methods.  Moreover, since the
self-training wrapper retains the scalability of the base method, we
obtain both higher quality and better scalability.
\end{abstract}

\maketitle 

  \section{Introduction}

%\paragraph*{put background on graph-based SSL and ``linear'' graph
%  diffusions, spectral and social}

  Graph data is prevalent and models entities (nodes) and the
  strong interactions between them (edges).   The source of these
  graphs can naturally come with the provided interactions (social
  networks,  views, likes, purchases, messages, links) or can be
  derived from metric data (embedded entities) by retaining only edges
  that correspond to closest neighbors.

  A fundamental task 
arises when 
label information is available only for a small set of  {\em
    seed} entities $(x_j,\vecy_j)$ $j\leq n_\ell$ and we are
  interested in learning labels for all other entities $x_i$ for $i\in
  (n_\ell,n_\ell+n_u]$.
See e.g. the surveys \cite{SemiSupervisedLearning:2006,GraphSSLbook:2014}.
The learning uses some smoothness assumption that similarity derived from the graph
  structure implies similarity of labels.
Often the graph structure is combined with other features by the
learning algorithms.

 Classic methods for semi-supervised learning and many other related 
 fundamental graph tasks (clustering, centrality, influence) are based on some 
 underlying diffusion process that propagates from a node or set of 
 nodes through the edges of the graph.  The diffusion defines 
dense fine affinity relations between nodes using the provided sparse 
set of strong interactions.  
With SSL, the affinity relation guides the label learning, for 
example, by a weighted aggregation of seed labels to obtain 
{\em soft labels}.

Most popular SSL methods can be interpreted through underlying
  {\em spectral} diffusions \cite{Chung:Book97a}, utilizing the graph Laplacian, 
graph cuts
  \cite{BlumChawla:ICML2001,BlumLRR:ICML2004}, and random walks.
They include label propagation \cite{ZhuGL:ICML2003}, 
label propagation using the normalized graph Laplacian
\cite{ZhouBLWS:NIPS2004,LinCohen:ICWSM2008}, and many 
variations.  The methods are highly scalable
and implemented 
using Jacobi iterations that 
roughly translate into repeated averaging over neighboring nodes.  The
algorithms are applied successfully 
to massive graphs with billions of edges \cite{RaviDiao:AISTATS2016}
using highly distributed platforms \cite{pregel:sigmod2010}.

 Another class of graph diffusions, which we refer to
 here as {\em social},  underline classic
social and economic models 
of centrality \cite{Bavelas:HumanOrg1948,Sabidussi:psychometrika1966,Freeman:sn1979,BlochJackson:2007,CoKa:jcss07,Opsahl:2010},
influence 
\cite{KKT:KDD2003,GRLK:KDD2010,DSGZ:nips2013,timedinfluence:2015}, and 
similarity \cite{CDFGGW:COSN2013} of nodes.  Social diffusion
propagate along  shortest-paths (distance 
 diffusion)  or reachability 
 searches (reach diffusion).
A powerful extension  defines a generative model from a graph by randomizing  the presence (with reach 
diffusion) or the length (with distance diffusion) of edges 
\cite{KKT:KDD2003,GRLK:KDD2010,CDFGGW:COSN2013,DSGZ:nips2013,timedinfluence:2015}
and then works with respective expectations.
Social diffusion, inspired by the independent cascade model of
\cite{KKT:KDD2003}  and the continuous time (distance-based) model of \cite{GRLK:KDD2010,CDFGGW:COSN2013,DSGZ:nips2013,timedinfluence:2015}
was
recently adapted to SSL \cite{semisupInf:2016}. The 
proposed algorithms scale very well: For the simpler
{\em nearest-seed} variant which 
matches each label to the closest seed node in each simulation of the 
randomized model we simply use small number of 
graph (Dijkstra) searches.
The use of distance or reachability sketching based on
\cite{ECohen6f} allows for highly scalable label learning also over
the sketched affinity matrix.

 Both spectral and social diffusion based SSL models scale well
even with a large number of labels, using heavy hitter sketches 
with label propagation \cite{RaviDiao:AISTATS2016} or 
naturally with social diffusion using sketches.
We interpret these methods as {\em linear} in the sense that
the diffusion propagates from seeds through edges
without amplifying strong signals or suppressing weak ones. 
% Moreover, seed nodes have a special role which can not be amplified
% by highly related nodes.

 Recently proposed  non-linear learning methods, based on
node embeddings and graph 
 convolutional networks,  had impressive success in improving the
 quality of the learned labels.
 In particular, DeepWalk~\cite{deepwalk:KDD2014} applied the
 hugely successful word embedding framework of \cite{Mikolov:NIPS13} to
 embed the graph nodes in a way that preserves the affinity relation
 defined by co-occurrence frequencies of pairs in short random walks:  A softmax 
  applied to inner products of embeddings approximates the frequency 
  of the pair. A supervised learning algorithm is then trained on the
  embedding vectors and labels of seed nodes.  Node2vec
  \cite{node2vec:kdd2016} refined the approach using hyperparameters
  that tune the depth and breadth of the random walks.   Another
  method, {\sc Planetoid}, used a multi-layer neural network instead of a
  single softmax layer \cite{Yang:ICML2016}.  With these methods, the
  lower-dimensional embeddings serve as a
 ``low rank''  representation of the affinity matrix  and the softmax
 and neural network introduce non-linearities that 
emphasize larger inner products.
 Another successful proposal are Graph convolutional networks (GCN)
 \cite{HenaffBL:2015,AtwoodT:NIPS2016,DefferrardBV:NIPS2016,KipfW:ICLR2017}, which
are neural networks with layers that
follow the graph adjacency structure.   Each layer applies a
 non-linear activation function, most often a sigmoid or ReLU, to the
 aggregate over neighbors. 

  This abundance of recent work introduced many new components and often at the
  same time:
  Low-rank embeddings, non-linear propagations, learning of weights of
  hidden layers (GCNs), learning of node
  weights~\cite{NandanwarM:kdd2016}.  
% An orthogonal component to all
%  methods is the integration of additional node features.
%  graph structure and combining them in the learning algorithm.
These recent methods, however, while demonstrating improved labeling quality,
do not scale as well as label propagation and methods based on
social graph diffusions.

  Our  aim here is to isolate the main contributor(s) to the
  quality gain of the recent non-linear approaches and to seek methods
that combine the scalability advantage of the simpler diffusion 
  models with state of the art labeling quality.

\ignore{
\edith{to edith:?  put that discussion later}\paragraph*{orthogonal ingredients to basic algorithms}
  The recent exciting new approaches have multiple ingredients making 
  it challenging to isolate the key ingredients to be credited for improvements 
  in prediction.
Another ingredient, inherent with  graph convolution
  networks \cite{HenaffBL:2015,DefferrardBV:NIPS2016,KipfW:ICLR2017}
  but can also be combined with other
  approaches \cite{NandanwarM:kdd2016},  is the learning of node or
  edge weights that minimize the empirical error.  The weights can be
  expressed as learned functions of other features.
 
 %   NN:  non linear intermediate aggregations,  random initializations
 %  of weights, learning via back propagation.  Many of the key
 %  benefits are due to the first two steps.   
}

% \paragraph*{Self training background}
 We explore placing these ``linear'' diffusions in a  {\em 
   self-training} ({\em bootstrapping}) framework.  Self-training is arguably 
 the earliest approach to SSL, dating back five decades to Scudder 
 \cite{Scudder1965}, and extensively studied 
by the NLP community
 \cite{Yarowsky:ACL1995,Abney:CL2004,WhitneySarkar:ACL2012}.
The self-training framework can be viewed as a wrapper around a {\em
  base} learning algorithm.  The base algorithm takes as 
input a set of labeled examples and makes predictions with associated 
margins or confidence scores for other examples.   The wrapper
applies the base algorithm in steps, where at the end of each step, the 
highest-confidence predictions are converted to become new labeled
examples.

Our {\em bootstrapped} diffusions retain the high
scalability of the base diffusions while introducing non-linearity
that allows us to work
with a richer class of models.
In particular, with our base algorithms, ``seed''  examples  have a special role
that is not amplified by  implied  high-confidence predictions.
Bootstrapping provides such amplification by promoting high-margin
predictions to ``seed'' roles.

% In a seemingly very different context,   Spielman and
% Teng~\cite{SpielmanTeng:sicomp2013} presented a local graph
% clustering algorithm that uses short random walks from seed nodes.  A
%property of their algorithm, introdced to control the running time
%bound,  is that it suppresses paths through
% less-visited nodes.  They provide a theoretical analysis showing that
% they can find low conductance cuts with guarantees next to the
% Cheeger inequality. 

  We perform experiments using linear diffusion models and their
  bootstrapped versions.  
We use classic 
Label propagation \cite{ZhuGL:ICML2003},  Label
  propagation using the normalized Laplacian \cite{ZhouBLWS:NIPS2004},
  and nearest-seed, which is the simplest 
  distance diffusion model~\cite{semisupInf:2016}.
We apply a very basic bootstrapping wrapper that works
  with a fixed fraction of highest-margin predictions in each step.
We focus on a multi-class setting, where each node is a member of one
class, even though most of the method can be extended to the
multi-label setting.

  We apply the different methods
to  benchmark data and seed sets 
  used and made available by previous 
  work~\cite{Yang:ICML2016,KipfW:ICLR2017}.  In particular, we use
  social, citation, and knowledge graph data sets.
We compare the quality of the learned labels to state of the art
baselines,  including {\sc DeepWalk}~\cite{deepwalk:KDD2014}, {\sc node2vec}~\cite{node2vec:kdd2016},
  {\sc Planetoid}~\cite{Yang:ICML2016}, and GCNs~\cite{KipfW:ICLR2017}.
  We also perform more elaborate experiments 
on additional data and seed sets and on the
well-studied planted
  partition  (stochastic block) model \cite{CondonKarp:2001} which is 
  often used to understand the performance of clustering and community
  detection algorithms.

Our main focus is the quality of
learning from the graph structure alone using diffused seed node
labels.  
  We observe that  bootstrapped diffusions consistently improved 
  the quality, by 1\% to 12\%,  over the  base diffusion, both for spectral and social 
  and across types of graphs.  
   The most surprising outcome was 
  that prediction quality on benchmark data exceeded that of all 
  recent non-linear methods.

The use of  additional available node (or edge) features can significantly increase labeling 
  quality but there are multiple ways to integrate them in the learning
  algorithm.    We consider the use of the raw provided node features
  or smoothing them through a graph diffusion.
 A simple supervised learning algorithm is then trained on the
(raw or diffused) feature vectors and class labels of seed nodes.  We applied this method
with and without bootstrapping.
 We  observed that both diffusion and bootstrapping significantly
 enhanced performance.  Furthermore, our results dominated
those reported (with the use of node
features) by the state of the art baselines 
{\sc Planetoid}~\cite{Yang:ICML2016} and GCNs~\cite{KipfW:ICLR2017}. 
In particular, GCNs lend themselves to a direct comparison as they can be viewed as a feature diffusion with
non-linearities applied after each set of edge traverasals and node/layer
weight tuned through back propagations.   It is interesting that we obtained comparable
or better results using bootstrapped linear models and without backprop training.

  The paper is organized as follows.  The linear base diffusion we use
  and the bootstrapping wrapper are discussed in
  Section~\ref{diffusions:sec}.  Our data sets and our experimental
  methodology are laid out in Section~\ref{experiments:sec}.   The
  results on the benchmark data are reported and discussed in
  Section~\ref{benchmark:sec}.  Additional experimental results  using
  additional data and seed sets are discussed in
  Section~\ref{randomsplits:sec}.  Detailed parameter study of the
  bootstrapping wrapper is provided in Section~\ref{moreexp:sec}.
  Section~\ref{features:sec} presents the experiments with feature vectors
  diffusions. We conclude in Section~\ref{conclu:sec}.

\section{Linear and Bootstrapped diffusions} \label{diffusions:sec}
\SetKw{nlabel}{label}
\SetKw{margin}{margin}

  In this section we review SSL methods based on linear (spectral and
  social) graph diffusions, while emphasizing the variants we 
  used in our experiments.
  In particular, we discuss two textbook
  label propagations methods \cite{ZhuGhahramani2002,ZhouBLWS:NIPS2004} and
  a distance-diffusion method \cite{semisupInf:2016}.
 We then discuss the application of self-training to these base methods.

\subsection{Spectral diffusion methods}
  The label vectors for seed nodes $\vecy_i$ are initialized so that
the entry that corresponds to the provided label $j\in [L]$ of each seed node $i\leq n_\ell$ is set to $y_{ij}=1$ and other entries $j'\not= j$ are set to $y_{ij'}=-1$.
The graph structure is represented by an affinity matrix $\boldW$, which can be provided as input or learned.
In our experiments, following prior work we compared with,
we used the adjacency matrix of the provided
undirected graphs with uniform edge weights and no self
  loops.  That is $W_{ij}=1$ when the edge $(i,j)$ is present and
$W_{ij}=0$ otherwise.  
% We will also use the diagonal degree matrix  $\boldD$, 
  % defined as $D_{ii} = \sum_j W_{ij}$ the raw sums of $\boldW$.
  
 The algorithms we use compute soft labels $\vecy_i$ for the unlabeled nodes $i>n_\ell$ of dimension that is equal to the number of classes $L$.
 The learned label we return for each node $i$
 follows the maximum entry $\arg\max_j y_{ij}$. We note
that often higher quality learned labels are obtaining by training a learning algorithm on the soft label \cite{semisupInf:2016} but in our experiments here we used these common simple class predictions.

\paragraph{{\sc Label propagation  (LP)} Zhu and Ghahramani \cite{ZhuGhahramani2002}}
Pseudocode is provided as Algorithm \ref{lp:alg} (As presented in \cite{SemiSupervisedLearning:2006}).
 A learned soft labels matrix $\boldY$ is initialized to the seed label vectors $\vecy_i$ for seed nodes $i\leq n_\ell$ and to $\vecy_i=\vec0$ for the unlabeled nodes $i>n_\ell$.
 The algorithm performs iterations where in each iteration,
 each non-seed node obtains a new soft label that is the weighted average of the soft labels from the previous iteration of its neighbors.
We note that this algorithm may not converge and does not necessarily improve with
iterations.  Therefore, in our experiments, we treat the number of iterations performed as a hyperparameter.

\begin{algorithm2e}[h]\caption{{\sc Label propagation (LP)} \cite{ZhuGhahramani2002} \label{lp:alg}}
  {\small
\SetKw{labels}{labels}
\DontPrintSemicolon
\KwIn{Affinity matrix $\boldW$. Provided labels $\vecy_1,\ldots \vecy_{n_\ell}$.}
\tcp{Initialization}
$\forall i,\ D_{ii}\gets \sum_j W_{ij}$\tcp*{diagonal degree matrix}
$\boldY^{(0)} \gets  (\vecy_1,\ldots,\vecy_{n_\ell},\vec0,\ldots,\vec0)$\;
\tcp{Iterate:}
\ForEach{$t=0,\ldots,T$}{
$\boldY^{(t+1)} \gets \boldD^{-1} \boldW Y^{(t)}$\;
$\boldY^{(t+1)}_{[n_\ell]\cdot} \gets
(\vecy_1,\ldots,\vecy_{n_\ell})$\tcp*{Reset learned labels of labeled
  nodes}
}
\tcp{Finalize:}
\ForEach(\tcp*[f]{Label points by largest entry}){$i > n_\ell$}{$\labels[i] \gets\arg\max_j y_{ij}$}
\Return $\labels$
}
\end{algorithm2e}

\paragraph{{\sc Normalized Laplacian label propagation} Zhou et al \cite{ZhouBLWS:NIPS2004}} Pseudocode is provided as Algorithm~\ref{ls:alg}.  This algorithm is related to Personalized Page Rank (PPR) and uses the normalized graph Laplacian \cite{Chung:Book97a}.  
%   This is essentially PPR variant.  The more standard random walk uses $D^{-1}W$ and we get 
% $$Y^{(t+1)}\gets \alpha (D^{-1}W) Y^{(t)} + (1-\alpha) Y^{(0)}$$
The soft labels $\boldY^{(t)}$ converge to $\boldY^{(\infty)}$
that satisfies
  \begin{equation} \label{limitlap:eq}
\boldY^{(\infty)} = \alpha (\boldI-(1-\alpha) \boldA)^{-1}
\boldY^{(0)}\ .
  \end{equation}
The soft labels at convergence $\boldY^{(\infty)}$ correspond to 
the stationary distribution when performing random walks from the seed nodes with some probability $\alpha$ of returning to the seed set in each step.  
Ideally, we would perform enough iterations so that the
learned soft labels $\boldY^{(t)}$ are close to $\boldY^{(\infty)}$.
With this algorithm,
the number of iterations is a parameter that trades off quality and computation, that is, we expect performance to improve with iterations.
The return probability $\alpha$ is a hyperparameter.

\begin{algorithm2e}[h]\caption{{\sc Normalized Laplacian LP}  \cite{ZhouBLWS:NIPS2004}\label{ls:alg}}
  {\small
    \SetKw{labels}{labels}
    \DontPrintSemicolon
\KwIn{Affinity matrix $\boldW$, provided labels $\vecy_1,\ldots
  \vecy_{n_\ell}$,  return probability $\alpha\in (0,1)$}
\tcp{Initialization:}
$\boldY^{(0)} \gets  (\vecy_1,\ldots,\vecy_{n_\ell},\vec0,\ldots,\vec0)$\;
$\forall i,\ \boldD_{ii}\gets \sum_j   W_{ij}$\tcp*{Diagonal degree matrix} 
$\boldA \gets \boldD^{-1/2} \boldW \boldD^{-1/2}$\tcp*{Normalized adjacency matrix}
\tcp{Iterate:}
\ForEach{$t=0,\ldots,T$}{
$\boldY^{(t+1)}\gets (1-\alpha) \boldA \boldY^{(t)} + \alpha  \boldY^{(0)}$\;}
\tcp{Finalize:}
\ForEach(\tcp*[f]{Label points by largest entry}){$i > n_\ell$}{$\labels[i] \gets\arg\max_j y_{ij}$}
\Return $\labels$
}
\end{algorithm2e}

\subsection{Social diffusion methods}
We consider recently proposed SSL methods based on distance diffusion \cite{SemiSupervisedLearning:2006}. 
The input is a 
directed graph $G=(V,E)$ where nodes $[n_\ell]$ are the seed nodes and a 
distribution $\distW$ that generates a set of lengths $\vecw > \vec0$ for the edges $e\in E$. 
The algorithm iterates the following.  It draws a set of edge lengths 
$\vecw^{(t)} \sim \distW$ from which it computes a set of soft labels $\vecy_i^{(t)}$ for $i>n_\ell$.  The soft label $\vecy_i$ is
computed from the shortest-path distances $d^{(t)}_{ij}$ from $i$ to each seed node $j\leq n_\ell$ and the respective labels $\nlabel[j]\in [L]$. 
The final soft label we seek for each $i>n_\ell$ is the expectation 
$\E[\vecy_i^{(t)}]$,  and we approximate it by the average over the 
iterations.  The number of iterations here is a parameter that 
trades off computation and quality. 

\paragraph{{\sc Nearest-seed} \cite{SemiSupervisedLearning:2006}}
The general formulation allows each soft label $\vecy_i^{(t)}$ to depend on the 
set of distances and labels $\{(d_{ij},\nlabel[j])\}$ of all seed nodes $j\leq n_\ell$ and requires distance sketching techniques to approximate using near-linear computation. 
In our experiments we focused only on the {\sc Nearest Seed} variant (pseudocode provided as Algorithm~\ref{nearestseed:alg}), where  in each iteration we only use the label of the seed node that is 
closest to $i$:
$$\vecy_i^{(t)} \gets \vecy_{\arg\min_{j\leq n_\ell}} d^{(t)}_{ij}\ .$$
The computation of each iteration (computing the closest seed to each node) 
is equivalent to one single source shortest path computation such as
(a slightly modified) Dijkstra's algorithm.  Hence it is near linear.

Since we use undirected graphs in our experiments, we generate two directed edges for each undirected edge.  Guided by \cite{semisupInf:2016} The lengths of edges are drawn independently from an exponential distribution with parameter that is equal to the inverse of the degree of the source node with possibly a fixed offset:
$$w(u,v) \sim \Exp[1/|\Gamma(u)|] + \Delta\ .$$  This achieves the effect that edges from nodes with larger degrees are in expectation longer and therefore less likely to participate in the shortest paths.  
The fixed offset that is added to edge lengths which allows as to control 
the balance between the number of hops and the weighted path length. 

We comment that our experiments did not exploit the full power of the rich class of distance-based model proposed in
\cite{semisupInf:2016}.  Our evaluation was limited to Nearest-seed and
exponentially distributed edge lengths with parameter equal to the
inverse degree.  The only hyperparameter we varied is  the offset $\Delta$.

\begin{algorithm2e}[h]\caption{{\sc Nearest Seed} \cite{semisupInf:2016}\label{nearestseed:alg}}
  {\small
%    \SetKw{labels}{labels}
    \SetKw{nearestS}{nearestS}
    \DontPrintSemicolon
    \KwIn{$G=(V,E)$, distribution $\distW$ of edge lengths, provided classes $\nlabel[i]\in [L]$ for nodes $i\leq n_\ell$}
\tcp{Initialize:}    
\ForEach{$i > n_\ell$}{$\vecc_i \gets \vec0$\tcp*{counter for nearest-seed label}}
\tcp{Iterate:}
\Repeat{$T$ times}{
Draw edge lengths $w \sim \distW$\;
\ForEach(\tcp*[f]{Apply a single-source shortest path algorithm on $G=(V,E,w)$ from seed nodes $[n_\ell]$ to compute the closest seed to each node $i$})
        {$i > n_\ell$}{
  $\nearestS[i] \gets  \arg\min_{j\leq n_\ell} d_{ij}$
}
\ForEach(\tcp*[f]{Label of the nearest seed}){$i > n_\ell$}{$c_{i, \nlabel[\nearestS[i]]} ++$}}
\tcp{Finalize:}
\ForEach(\tcp*[f]{Label points by largest entry}){$i > n_\ell$}{$\nlabel[i] \gets\arg\max_j c_{ij}$}
\Return $\nlabel$
}
\end{algorithm2e}

\subsection{Bootstrapping Wrapper}
The self-training framework has many variants~\cite{Yarowsky:ACL1995,Abney:CL2004}.
The bootstrapping wrapper takes as input a base algorithm $\calA$ and a set $S$ of labeled seed nodes.  In each step the algorithm $\calA$ applied to $S$ and returns a set of learned labels $\nlabel$ for all nodes not in $S$ together with prediction margins $\margin$.  The wrapper then
augments the set $S$ with new nodes.
In our experiments we used a very basic formulation, shown as Algorithm~\ref{bootstrap:alg}.
We  fix the
number of new seed nodes from each class to be selected at each step to be
proportional to the respective frequency $\pi_i$
of the class $i$ in the data, using a proportion parameter $r$.
More precisely, at each step $t$,  we consider for each class $i$, the set of
all non-seed nodes with learned labels  $\nlabel^{(t)}[j]$ ($j\not\in S$) in the class $i$.  We then take
the $r \pi_i$ nodes with highest-margin learned labels as new seeds added to $S$.  If there are fewer than $r \pi_i$ nodes with learned label $i$, we take all these nodes to be new seeds.
We expect quality to decrease with $r$ but also that the number of steps needed to
maximize the bootstrapped performance to decrease with $r$. 
The algorithm terminates when all nodes become seed nodes but each step $t$ provides a full set of learned labels $\nlabel^{(t)}$ for all nodes.
The precision may first increase with the step $t$  but then might
decrease  and we therefore  use cross validation or a separate
validation set to determine which set of learned labels
$\nlabel^{(t)}$ to use.   Validation can also be used to stop the
algorithm when precision starts dropping.
The parameters we use here are  $r$, which trades off computation and
quality and the hyper/parameters of the base algorithm $\calA$.

\begin{algorithm2e}[h]\caption{Basic {\sc Bootstrapping Wrapper} \label{bootstrap:alg}}
  {\small
%    \SetKw{labels}{labels}
    \DontPrintSemicolon
    \KwIn{Seed set $S \subset V$ with labels $\nlabel:S\in [L]$, $r$, frequencies $\pi_i$ for $i\in[L]$,  Base algorithm $\calA$ that given $S$ and $\nlabel:S$ augments $\nlabel$ to $V\setminus S$ and provides $\margin:V\setminus S$.}
%    \tcp{Initialize:}
    $t \gets 0$\;
  % \tcp{Iterate:}
\Repeat(\tcp*[f]{Main iteration})
       {$S==V$ }{
         Apply $\calA$ to seeds $S$ and  $\nlabel:S$ to
         assign $\nlabel: V\setminus S$\;
         $\nlabel^{(t)} \gets \nlabel$\tcp*{Remember step $t$
           predictions}  t++\;
         \ForEach{class $i\in L$}{
         $C \gets \{j \in V\setminus S \text{ such that } \nlabel[j] == i\}$\;
           Place in $S$ the $\min\{|C|,r\pi_i\}$  highest $\margin$ nodes in $C$}
}
\tcp{Finalize:}
\Return $\nlabel^{(t)}$ that is best on validation set
}
\end{algorithm2e}

\ignore{
\paragraph{Spielman Teng Lazy walk}

  For an unweighted undirected graph we consider the adjacency matrix
  $A$ (with self loops) and diagonal degree matrix $D^{-1}$.  The self
  loop originate from Lovasz-Simonovitz and are needed to smooth the
  walk for convergence proofs.     We consider the ``lazy  walk''
  matrix $M = (AD^{-1}+I)/2$ (probability half of ``staying'',
  probability half of choosing neighbor).

  For the purposes of local clustering, we are only interested in few
  steps of the walk, before it converges to the stationary
  distribution. 

   We initialize a vector $q_0$ with the seed set $S$.

  The base diffusion process sets $q_i \gets Mq_{i-1}$.

  Here we do not seek convergence, as this will not depend on the
  start point.  We just do that for few iterations.

  The bootstrapped version with nonlinearity  uses few iterations, 
suppresses (rounds to $0$) small entries of $q$ and may increase
larger entries (say normalized power or softmax).  It then continues
with the new vector.

}

% \paragraph{Kipf and Welling}
%  Base diffusion:  Weighted averaging of self and neighbors.
%   KW add randomization and nonlinearity.  Essentially the RELU passes
%  on ``variance''  (larger positive contributions) which is higher
%  when value is higher.

% \input{experiments.tex}
\section{Datasets and experimental setup}  \label{experiments:sec}
To facilitate comparison, we use the benchmark dataset and seed set
combinations used in prior work \cite{Yang:ICML2016}.
In our detailed experiments, we use additional data sets, multiple seed sets, and synthetic data sets.

We limited our evaluation to data sets that are multi-class but not
multi-label, meaning that each node is a member of exactly one class.
We note that the base and bootstrapped algorithms we use can naturally be extended to a multi-label
setting but there are different mechanisms to do so
\cite{node2vec:kdd2016,semisupInf:2016} that may or may not assume
that the number of classes of each node is provided. 
Moreover, some of the algorithms we compare with \cite{KipfW:ICLR2017} do not have multi-label variants.
We therefore opted to remove this variability by only considering multi-class.

Table~\ref{data:tab} summarizes the statistics of the datasets we
used.   For each dataset we report the number of nodes, edges and
classes.  We also list the labeling rate, which is the fraction of the
nodes we used as seed labels.  The data sets include three citation
networks:  {\sc Cora}, {\sc Pubmed}, and {\sc Citeseer} from
\cite{Sen:AI2008},  one
Knowledge graph (entity classification) dataset ({\sc NELL})
preprocessed by \cite{Yang:ICML2016} from \cite{Carlson:AAAI2010}, and the
{\sc YouTube} group membership data set from
\cite{MisloveGDB:IMC2007}.  We preprocessed the YouTube data by removing groups
with less than 500 members and not using for training or testing the
users that were members of more than one group.  This left us with
14355 labeled nodes to use as seed/validations/test sets.
% as reported in~\cite{Yang:ICML2016}.

Our seed set selection followed \cite{Yang:ICML2016}.
For the citation networks, the seed sets 
 contain 20 nodes randomly selected from each class.  
 For the NELL data, with 210 classes, the seeds selection was proportional to class size % (and at least 1) 
 with labeling rates $0.1$, $0.01$ and $0.001$.  For the YouTube data 
 we used 50 seeds from each of the 14 classes. 

The citation and knowledge graph 
datasets have associated bag-of-words node features which can be used
to enhance prediction quality.  We report experiments that use these
node features in Section~\ref{features:sec}.

%  in multiple way.  As we explained in the
% introduction, we opted to remove this variability and only use the (undirected) graph structure for label learning.

%The benchmarks provided in \cite{Yang:ICML2016} specified seed set and test set selection, which we follow to facilitate a comparison.
%We also perform a separate detailed evaluation over multiple random splits of each dataset.

We also use synthetic data generated using the planted partition
 (stochastic block) random graph model \cite{CondonKarp:2001}.  These random graphs are specified by a number
$L$ of equal-size classes, the total number of nodes, and two parameters
$q<p \in [0,1]$.   The graph is generated by instantiating each
intra-class edge independently with probability $p$ and each
inter-class edge independently with
probability $q$.  The sets of parameters we used to generate graphs
are listed in Table~\ref{datarandom:tab}.   Our seed sets had an
equal number of randomly selected nodes from each class.

\begin{table} \caption{Dataset statistics \label{data:tab}}
{\small
\center
\begin{tabular}{c|rrrr}
Dataset & \#Nodes & \#Edges & \#Classes & label rate \\
\hline
{\sc Citeseer} & 3,327 & 4,732  & 6 & 0.036\\
{\sc Cora} & 2,708 & 5,429 & 7 & 0.052\\
{\sc Pubmed} & 19,717 & 44,338 &  3  & 0.003\\
{\sc NELL} & 65,755 & 266,144 & 210 & 0.1,0.01,0.001\\
{\sc YouTube} & 1,138,499 & 2,990,443 & 14  & 0.00023 
\end{tabular}
}
\end{table}

\begin{table} \caption{Planted partition datasets \label{datarandom:tab}}
{\small
\center
\begin{tabular}{c|rrrrr}
  Model & \#nodes & \#Classes & $p$ & $q$ & label rate \\
  \hline
{\sc Planted Partition 3} & 3000  & 3 & 0.02 & 0.01 & 0.10 \\
{\sc Planted Partition 5} & 5000 & 5 &  0.018 & 0.01 &  0.10 \\
% {\sc Planted Partition 5a} & 5000 & 5 &  0.03 & 0.01 &  0.06 \\
{\sc Planted Partition 10} & 5000 & 10 & 0.025  & 0.01  & 0.20  
\end{tabular}
}
\end{table}

% cat *AB.csv |  gawk 'BEGIN{FS=",";}{print $2 " " $NF " " $3 " " $4 }' | sort

\begin{table*}[!ht]  \caption{Results on benchmark data \label{comparison:tab}}
\center 
{\small 
\begin{tabular}{c|llllll}
Method & {\sc Citeseer} & {\sc Cora} & {\sc Pubmed} & {\sc NELL 0.1} & NELL 0.01 & NELL 0.001 \\
\hline 
{\sc TSVM} \cite{Joachims:ICML1999} *\cite{Yang:ICML2016} & {\bf 0.640} & 0.575 &  0.622 & & & \\
{\sc DeepWalk} \cite{deepwalk:KDD2014} *\cite{Yang:ICML2016} & 0.432 &
                                                                       0.672 & 0.653 & 0.795 & 0.725 & 0.581 \\
{\sc DeepWalk} \cite{deepwalk:KDD2014} & 0.511 & 0.724 & 0.713 & 0.848 
                                                                     &
                                                                       0.748& 0.665 
  \\
{\sc node2vec} \cite{node2vec:kdd2016} & 0.547 & 0.749 &  0.753 &
                                                                  0.854 & 0.771 &0.671 
  \\
{\sc Planetoid-G} \cite{Yang:ICML2016} *\cite{Yang:ICML2016} & 0.493 &
                                                                       0.691 
                                     & 0.664 & 0.845 & 0.757 & 0.619 
  \\
{\sc Graph Conv Nets} \cite{KipfW:ICLR2017} &  &  &  & & & 0.660 \\
\hline\hline 
{\sc Nearest-seed} \cite{semisupInf:2016} & 0.490  & 0.710  & 0.751 & 0.859 & 0.793 & 0.700 \\
\textcolor{blue}{+Bootstrapped} & 0.511 & 0.762 &  0.757 &  {\bf 0.860} & 0.801  &
                                                                 0.757\\
\hline 
{\sc Label Propagation}\cite{ZhuGhahramani2002} & 0.518  & 0.717 & 0.725 & 0.827  &
                                                                     0.775  & 0.600 \\
\textcolor{blue}{+Bootstrapped}& 0.533 & 0.780 & 0.749 & 0.849 & 0.818  & 0.731 \\
\hline 
{\sc Norm Lap LP} \cite{ZhouBLWS:NIPS2004} & 0.514 & 0.720 & 0.721 &
                                                                  0.840 & 0.790 & 0.667 \\
\textcolor{blue}{+Bootstrapped} & 0.536 & {\bf 0.784} & {\bf 0.788} &
                                                                      0.842 & {\bf 0.829} & {\bf 0.785}

\end{tabular}
}
\end{table*}

   We used our own Python implementation of the three base linear diffusions discussed in Section~\ref{experiments:sec}:
{\sc Label propagation (LP)} \cite{ZhuGhahramani2002} (Algorithm~\ref{lp:alg})
{\sc Normalized Laplacian LP} \cite{ZhouBLWS:NIPS2004} (Algorithm~\ref{ls:alg})
and {\sc Nearest Seed} \cite{semisupInf:2016} (Algorithm~\ref{nearestseed:alg}) and 
also the bootstrapping wrapper (Algorithm~\ref{bootstrap:alg}).

Our experimental study has multiple aims.   The first is to understand the
 attainable precision by the different algorithms, base and
 bootstrapped, and compared it to baseline methods.   For this purpose we 
used a wide range of hyperparameters with a validation
 set to prevent overfitting.  The second is to perform a parameter
 study of the bootstrapped methods.  The third focuses on scalability
 and considers results within a specified computation budget.

   We applied the following hyper/parameters.  
We run all bootstrapped algorithms with $r\in\{0.02,0.03,0.04,0.1,0.15,0.2,0.25,0.3\}$
as the fraction of nodes selected as seeds in each step (see Algorithm~\ref{bootstrap:alg}).
For bootstrapped LP  and bootstrapped normalized Laplacian LP we used $\{10, 20, 40, 100\}$ iterations of the
base LP algorithm in each step.  For the nearest-seed we used
$\{25,75,100,400\}$ iterations in each step.
For the normalized Laplacian we used return probabilities
$\{0.0001,0.01,0.05,0.1,0.2,0.5\}$ and for the nearest-seed we used
$\Delta\in\{1,10,50,100\}$.
In 
retrospect, all our offset choices for nearest-seed had similar
performance results.  
The normalized Laplacian LP consistently performed best with return 
probabilities in the range $\{0.1,0.2\}$.
As for the base algorithms performance.  For LP we
used a validation set to select the iteration to use (as
discussed in Section~\ref{diffusions:sec}) whereas with the normalized
Laplacian LP and nearest-seed we took the last iteration.

\section{Results on benchmark datasets}  \label{benchmark:sec}

In this set of experiments we follow the benchmark
seed set and test set selection of
  \cite{Yang:ICML2016,KipfW:ICLR2017} with the properties as  listed in Table~\ref{data:tab}.

The provided test sets for the citation networks 
included exactly 1000 randomly selected nodes. The tests sets for the
NELL data were slightly smaller.
Our evaluation in this section is intended to study the attainable quality by the different methods and we therefore use a 
 generous range of hyperparameters. 
An important issue was that a separate validation set  was not available with the benchmark data and moreover, because of unknown mappings, we could not
  produce one from the original raw data without overlapping with the
  provided seed and test sets.  To facilitate a fair comparison that provides learned labels for all test set nodes
  without overfitting (for all methods with hyperparameters), we did the following:  We randomly
 partitioned each provided test set to two equal parts $A$,$B$ and performed two
  sets of executions: In one set $A$ was used as a validation set for
  hyperparameter selection and $B$ for  testing and vice versa. 
% We note here
%  that we also experimented on different splits, where 120 nodes were
%  used for validation and the rest for prediction, and obtained
%  consistent results.   

The results are reported in Table~\ref{comparison:tab}  for the three linear
diffusion methods, their bootstrapped variants, and the following
baseline methods:
\begin{itemize}
\item
{\sc node2vec} \cite{node2vec:kdd2016}:  We used the published Python
code of \cite{node2vec:kdd2016} (which
builds on~\cite{Mikolov:NIPS13,deepwalk:KDD2014}) to
  compute an embedding.  We then applied logistic regression trained
  on the embeddings and labels of the seed to learn labels for other
  nodes.  The method uses two hyperparameters $(p,q)$ to
  tradeoff the depth and breadth components of the random walks.
 We used all combinations of the values $\{0.25,0.5,1.0,2.0,4.0\}$ for
 $p,q$ and set other hyperparameters (10 epochs, 10
  walks, walk length 80, embedding dimension 128) as in
  \cite{node2vec:kdd2016} .  We use the same validation/test
  splits to prevent overfitting.
\item
{\sc DeepWalk}~\cite{deepwalk:KDD2014}:  
The parameter setting $p=q=1.0$ with {\sc node2vec} implements {\sc DeepWalk}.  We
  list the results we obtained in this implementation.
%   Since the hyperparameter search was
%  not applied for deepwalk, we simply list the results on the original
%  test set.   
For reference, we also list the results as reported in
  \cite{Yang:ICML2016} using a different implementation.
\item
 {\sc Planetoid-G} \cite{Yang:ICML2016}.  Planetoid uses embeddings
 obtained using a multi-layer neural network, building on the graph
 relations captured by DeepWalk.  Planetoid-G is the
  (transductive) variant that does not use the node features.  We
  report the results as listed in \cite{Yang:ICML2016}.
\item
    Transductive support vector machines ({\sc TSVM})
  \cite{Joachims:ICML1999}, for which we list the results reported in \cite{Yang:ICML2016} when available -- due to scalability issues.
\item
Graph convolutional networks ({\sc GCN}):  Kipf and Welling~\cite{KipfW:ICLR2017}
used the benchmark data and seed sets of \cite{Yang:ICML2016} but only reported
results with the use of node features.  We only list their results on
the NELL dataset, where node features did not enhance
performance according to \cite{Yang:ICML2016}.
\end{itemize}

  We can see that the bootstrapped variants 
  consistently improve over the base methods.  We also observe that 
  (except in one case),  the best result  over all methods, including
  the baselines,  is 
  obtained by a bootstrapped diffusion method.  We can also observe
  that the base linear methods, the classic spectral and the recent
  social, perform competitively to state of the art and the
  bootstrapping elevates performance to above it.
We note that the results for LP reported in \cite{Yang:ICML2016} used a different
implementation (Junto with a fixed number of 100 iterations).

%\edith{Would be nice to have if we could: Run Kipf Welling without 
%  node features to benchmark experiments}

\begin{table*}[!ht]  \caption{Results averaged over random splits (seed set selections) \label{bootstrapped:tab}}
\center 
{\small
\begin{tabular}{c|lll l l |lll}
Method & {\sc Citeseer} & {\sc Cora} & {\sc Pubmed} & {\sc NELL}  &
                                                                    {\sc
                                                                    YouTube}
  & \multicolumn{3}{c}{{\sc planted partition}} \\
    &  &  &  &  0.001 &  &$L=3$  & $L=5$ & $L=10$ \\
$\times$ repetitions   & 100  & 100  & 100  &  10 & 5  & 10 & 10 & 10  \\
\hline\hline
{\sc DeepWalk} \cite{deepwalk:KDD2014} &  0.472 & 0.702  & 0.720  & 0.652 & & & &\\
{\sc node2vec} \cite{node2vec:kdd2016} & 0.473  & 0.729 & 0.724  &
                                                                   0.652 & & & &\\
\hline\hline 
{\sc Nearest-seed} \cite{semisupInf:2016}  & 0.470  & 0.717 & 0.726  &
                                                                  0.686
                                                                  &
                                                                    0.363
&  0.541 & 0.411  & 0.358 \\
\textcolor{blue}{+Bootstrapped} & 0.480 & 0.746 & 0.748  & 0.763  &
                                                                    {\bf
                                                                    0.512}
&  0.654  & 0.593 &  0.474\\
\hline
{\sc Label Propagation}\cite{ZhuGhahramani2002} & 0.479  & 0.728  &
                                                                    0.709
                                                    & 0.598 & 0.251&
                                                                     0.593 &0.513 &
                                                                      0.488\\
\textcolor{blue}{+Bootstrapped} & 0.496 & 0.781 & 0.747  & 0.690 &
                                                                   0.411
  & {\bf 0.753}  & {\bf 0.774}  & {\bf 0.657} \\
\hline
{\sc Norm Lap LP} \cite{ZhouBLWS:NIPS2004}  & 0.490 & 0.730  & 0.739
                                                    & 0.673 & 0.293 &
   0.592  & 0.507
      &  0.474 \\
\textcolor{blue}{+Bootstrapped} & {\bf 0.503} & {\bf 0.782} & {\bf 0.756}  & {\bf 0.791}  &
                                                                    0.431
  & 0.711 & 0.661   & 0.535
\end{tabular}
}
\end{table*}

\section{Random splits experiments} \label{randomsplits:sec}
We performed additional experiments in order to obtain a more robust
picture on the bootstrapping performance gain. For each data set we
performed multiple repetitions using different random splits of
 the data to seed, test, and validation sets  and averaged the results.

We used the five real-word datasets listed in Table~\ref{data:tab}.
For the citation networks ({\sc Citeseer}, {\sc Cora}, and
{\sc Pubmed}) we
used the raw data that included the full node labeling.
We used labeling rates as listed, with balanced seed sets for 
the citation networks and {\sc YouTube}.
 For the NELL 
graph, produced by~\cite{Yang:ICML2016}, we only used the provided 
2000 labeled nodes and selected a seed set with 0.001 labeling rate.
We used three {\sc planted partition} parameters as listed in
Table~\ref{datarandom:tab}.  A new planted partition graph was generated according
to these parameters for each repetition.
With all experiments we used a separate validation set of size 500
and used all remaining labeled nodes as a test set.
The average precision results are reported  in 
Table~\ref{bootstrapped:tab}.  The table also lists  the number of 
repetitions we used for each dataset.

%  We used the same seed set structure as in the benchmark 
%data, validation sets of size 500 and generally used all remaining 
%labeled nodes as a test set.  For youtube data we used 50 seeds from 
%each of the 14 classes.   

We also report results for the {\sc node2vec} and {\sc
  deepwalk} methods, when the computation of the embeddings was 
feasible for us, using the in-memory Python implementation
\cite{node2vec:kdd2016}.  Results are provided for the citation networks 
and NELL datasets.  We had to exclude the much larger YouTube data set and also 
the planted partition, since we generated a new graph 
for each repetition which would require a new embedding (a 
single embedding sufficed for all repetitions on the real-word networks).
Note that {\sc node2vec} was the top performer 
among the baseline methods on the benchmark data so it provides a good
reference point.

 We can see that the results confirm our observations from the benchmark data experiments:
 The bootstrapped methods consistently improve over
 the respective base methods and also consistently achieve the best results.  
The best performer is most often the bootstrapped normalized
 laplacian LP.  We suspect it is due to the flexibility provided
 through the return probability hyper parameter which controls the
 depth versus breadth of the propagation.    Such flexibility is
 provided in part by the iterations parameter of the basic label
 propagation method  and also by the 
hyperparameters of the distance diffusion
 models.  For the latter, we did not use incorporate this flexibility
 in our experiments.  
We observe that the two spectral methods perform similarity whereas the social method
seems to supplement the spectral methods and perform well on different data sets.  

  Our experiments with the planted partition random graphs also
  demonstrate a clear bootstrapping gain in quality with respect to the baselines.
This generative model is used extensively in experiments and analysis
of clustering algorithms.  
%Most
%often, using spectral methods such as personalized page rank (that is
%highly related to the normalized laplacian label propagation).   
The consistent precision gain by bootstrapping suggests wider
applicability and seeking better theoretical understanding of the
limits of the approach.

Finally, we note that we tested the bootstrapping wrapper with two other
implementations
of Label Propagation including: The sklearn Python library and
  Junto \cite{ZhuGhahramani2002}\footnote{\url{https://github.com/parthatalukdar/junto}} and observed similar performance gains over the base methods.
  % label   propagation implementations (Junto uses label propagation
  % \cite{ZhuGhahramani2002} with a different normalization after each step).

\begin{figure*}
\center 
\notinarxive{
\includegraphics[width=0.22\textwidth]{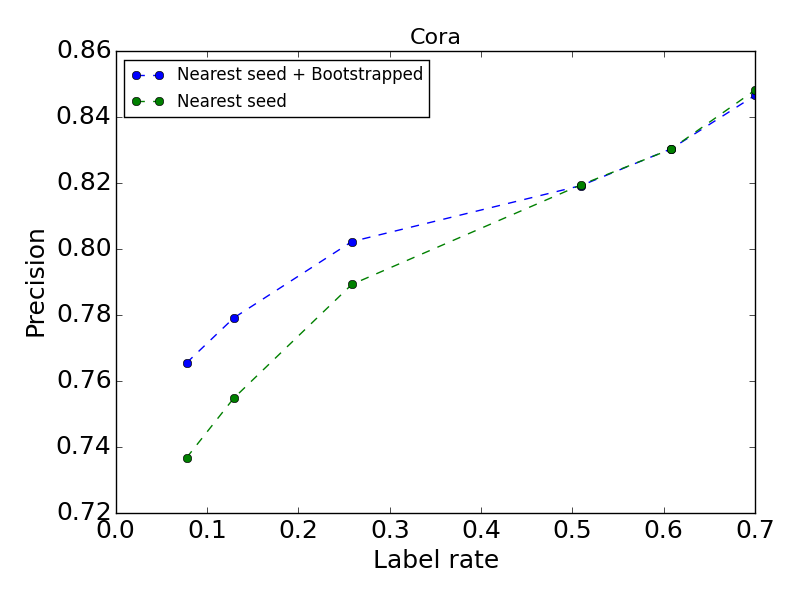}
\includegraphics[width=0.22\textwidth]{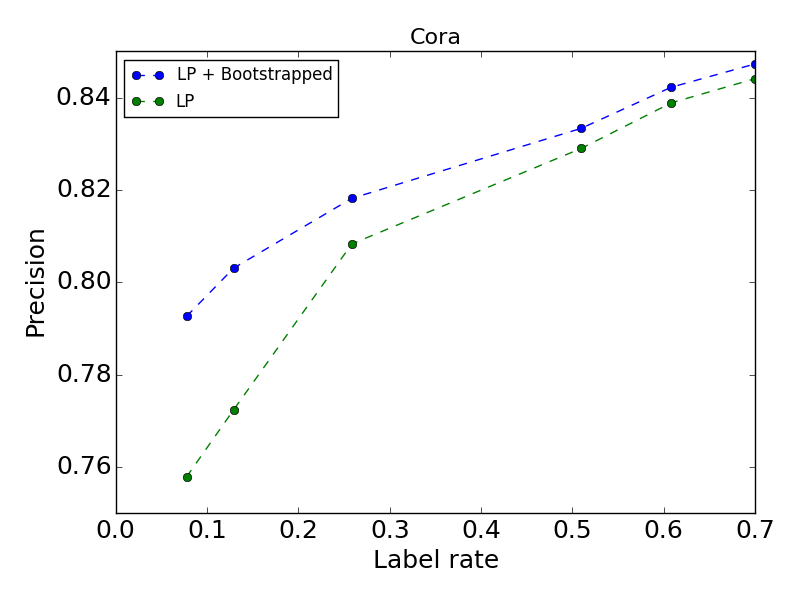}
\includegraphics[width=0.22\textwidth]{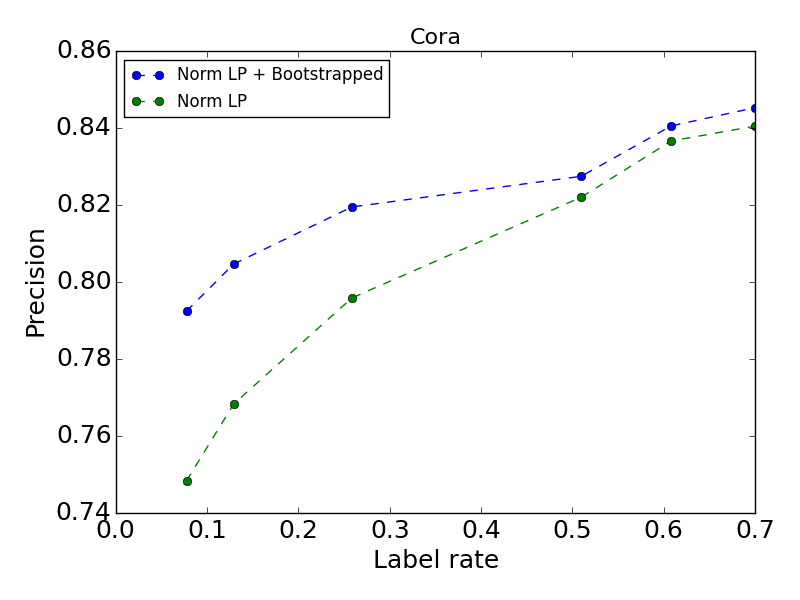}
\includegraphics[width=0.22\textwidth]{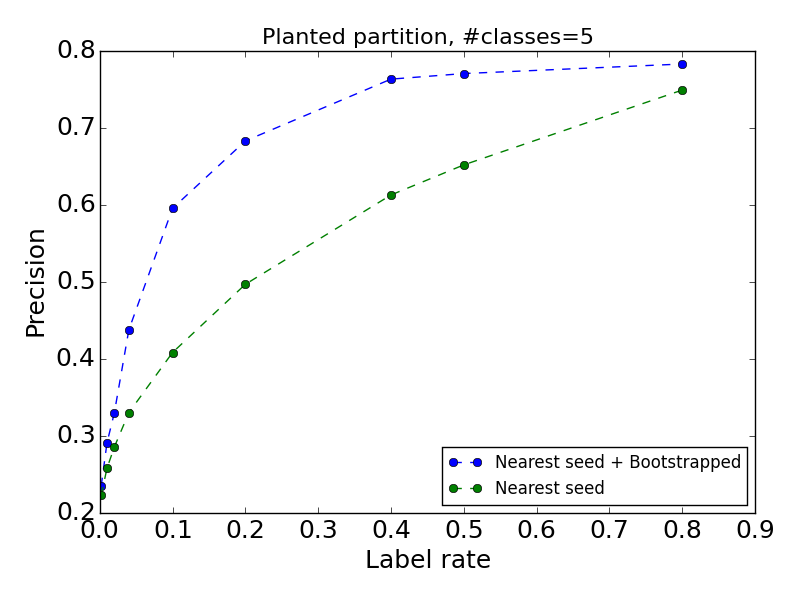}
\includegraphics[width=0.22\textwidth]{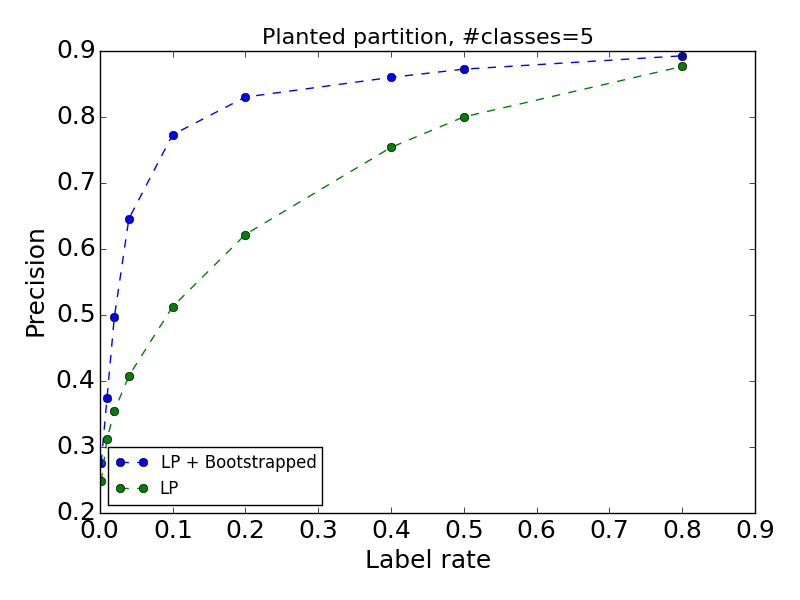}
\includegraphics[width=0.22\textwidth]{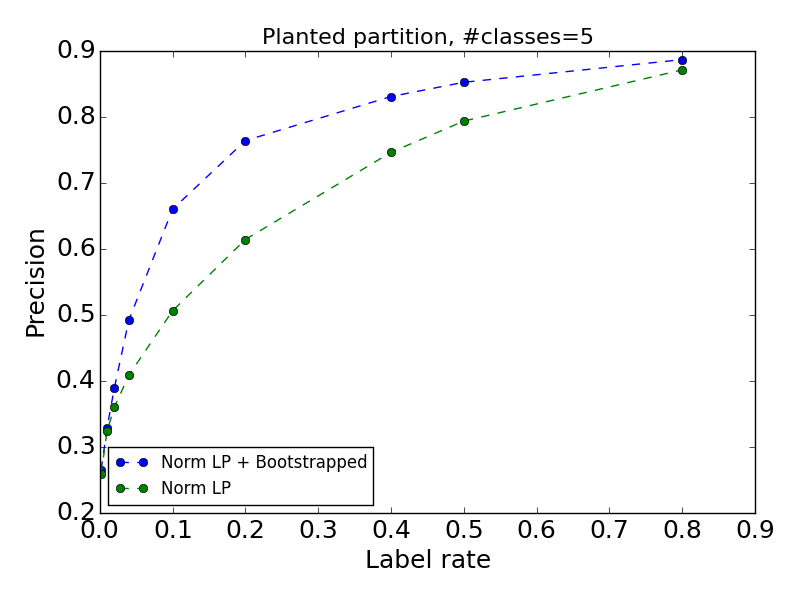}
\includegraphics[width=0.22\textwidth]{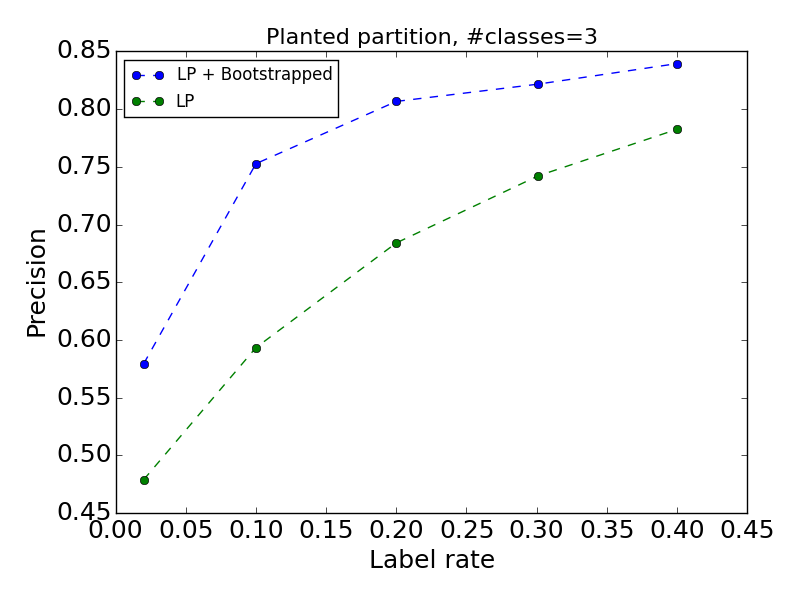}
\includegraphics[width=0.22\textwidth]{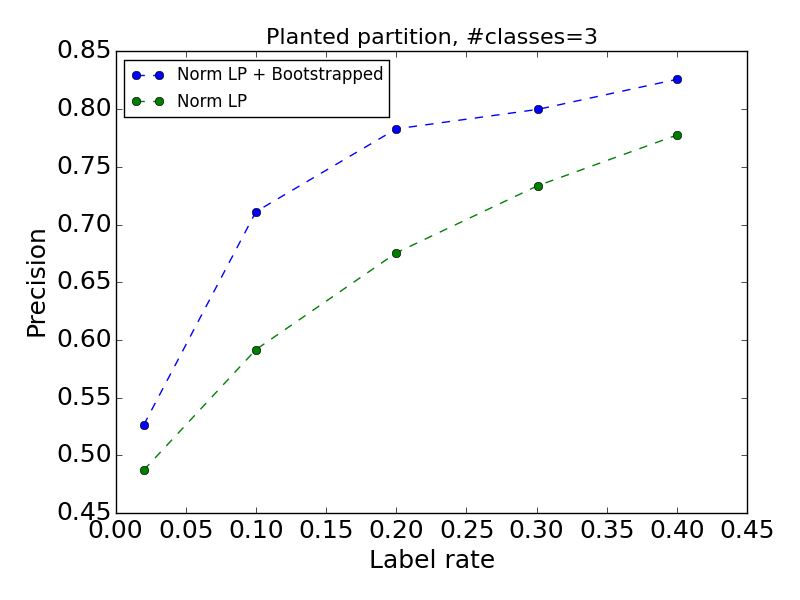}
}
\onlyinarxive{
\includegraphics[width=0.22\textwidth]{cora_graph_NN_seed_size_vs_accuracy.png}
\includegraphics[width=0.22\textwidth]{cora_graph_LP_seed_size_vs_accuracy.png}
\includegraphics[width=0.22\textwidth]{cora_graph_NormLP_seed_size_vs_accuracy.png}
\includegraphics[width=0.22\textwidth]{random_graph_0_NN_seed_size_vs_accuracy.png}
\includegraphics[width=0.22\textwidth]{random_graph_0_LP_seed_size_vs_accuracy.png}
\includegraphics[width=0.22\textwidth]{random_graph_0_normLP_seed_size_vs_accuracy.png}
\includegraphics[width=0.22\textwidth]{random_graph_1_LP_seed_size_vs_accuracy.png}
\includegraphics[width=0.22\textwidth]{random_graph_1_normLP_seed_size_vs_accuracy.png}
}
\caption{{\small Precision for varying labeling rates for base and bootstrapped 
        LP, Normalized Laplacian LP, and nearest-seed.  On selected datasets.}}
\label{precisionVSlabelingrate:fig}
\end{figure*}

\begin{figure*}
\center 
\notinarxive{
\includegraphics[width=0.32\textwidth]{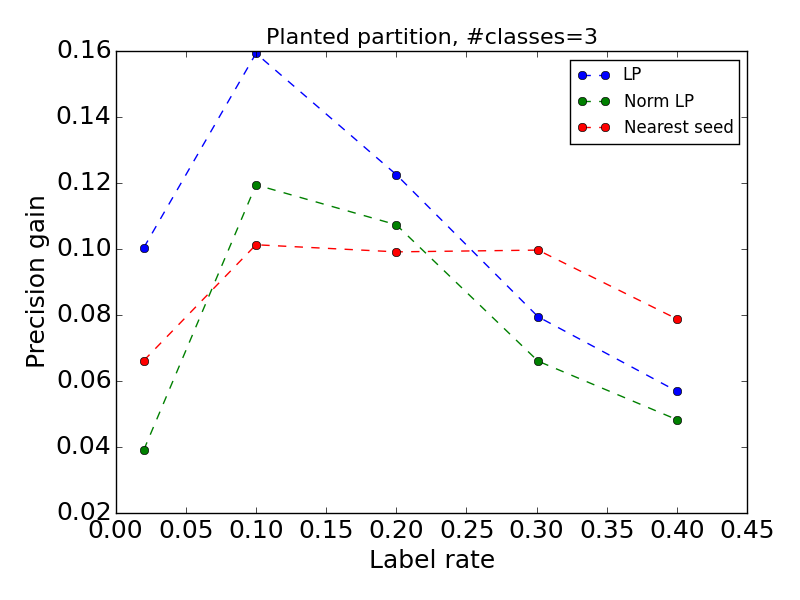}
\includegraphics[width=0.32\textwidth]{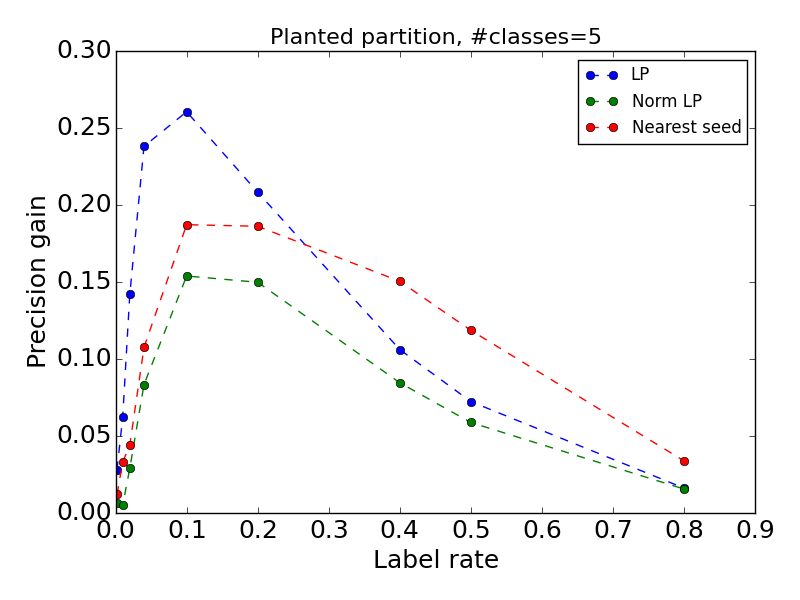}
\includegraphics[width=0.32\textwidth]{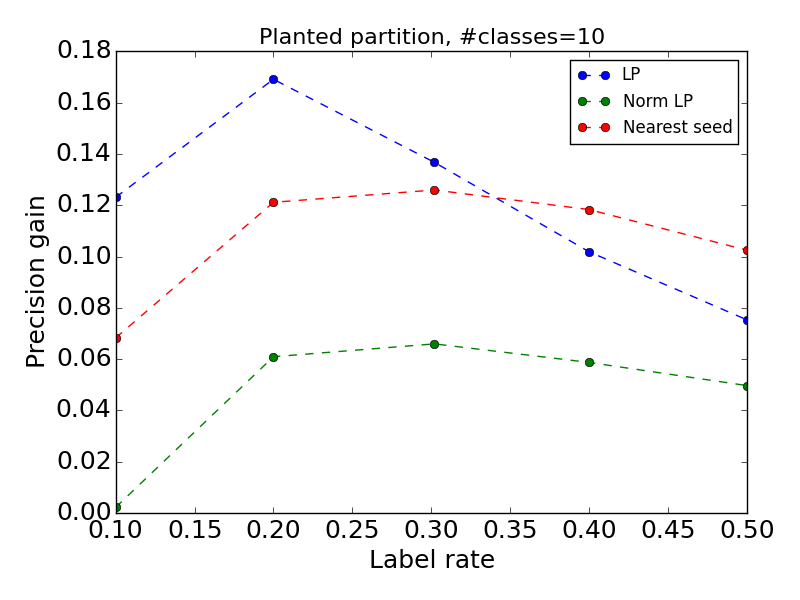}
}
\onlyinarxive{
\includegraphics[width=0.32\textwidth]{random_graph_1_label_rate_vs_precision_gain.png}
\includegraphics[width=0.32\textwidth]{random_graph_0_label_rate_vs_precision_gain.png}
\includegraphics[width=0.32\textwidth]{random_graph_2_label_rate_vs_precision_gain.png}
}
\caption{{\small Precision gain of bootstrapping for varying labeling 
    rates for     LP, Normalized Laplacian LP, and nearest seed.}}
\label{precisionVSlabelingrateGain:fig}
\end{figure*}

\section{Bootstrapping parameter study} \label{moreexp:sec}

 In this section we take a closer look and study how the
 bootstrapping quality gain depends on properties of the data and parameters.

\subsection{Labeling rate}
   We study the precision of both the base and bootstrapped algorithms as
   a function of the labeling rate.   Here we used $10\times$ random
   splits of the data sets.
   For the citation networks and planted partition graphs,
we varied the labeling rate while maintaining balanced seed sets that have the same number of seeds from each class.
   Representative results 
   showing the precision as a function of the labeling rate for selected data sets are visualized in Figure~\ref{precisionVSlabelingrate:fig}.
Figure~\ref{precisionVSlabelingrateGain:fig} shows the
gain in precision due to bootstrapping over the respective base method,
as a function of the labeling rate.

The plots show that, as expected, precision of all methods increases with
the labeling rate and that bootstrapping consistently improves performance.
We can see that across methods,
the gain in precision due to bootstrapping is smaller at the extremes,
when the labeling rate and precision are very low or very high.  The largest gains are obtained in the middle range.

\subsection{Seed set augmentations}
\begin{figure}
  \center
\notinarxive{
  \includegraphics[width=0.22\textwidth]{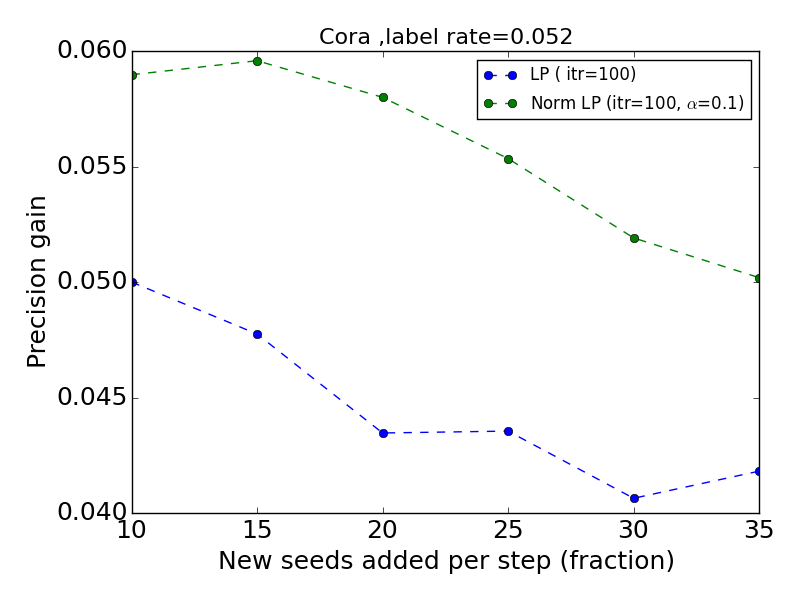}
  \includegraphics[width=0.22\textwidth]{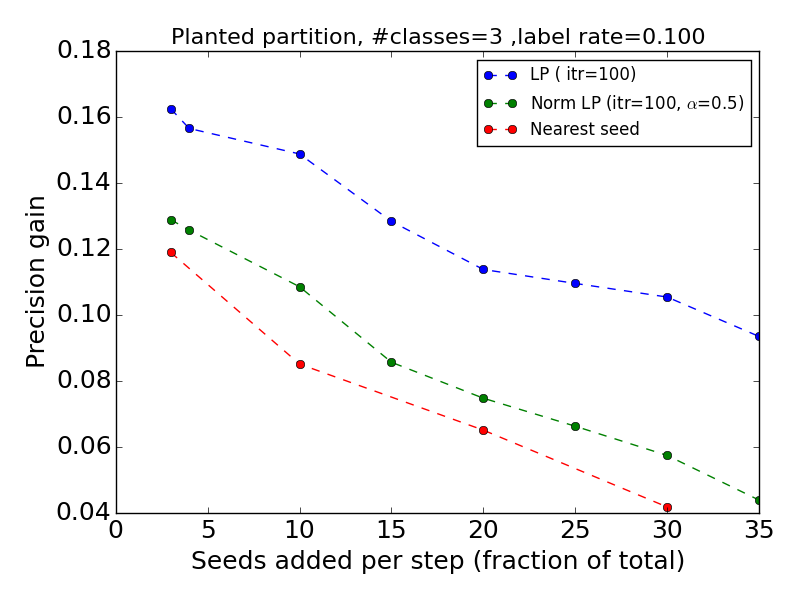}
  \includegraphics[width=0.22\textwidth]{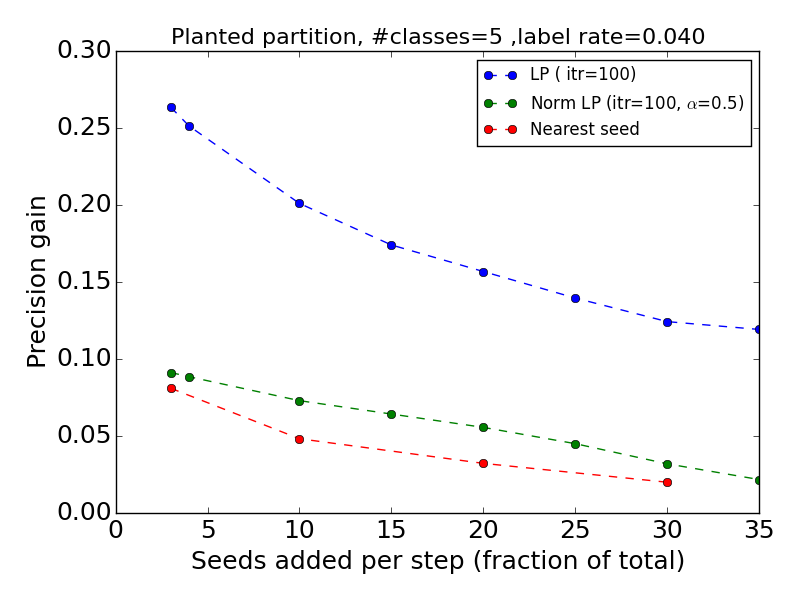}
    \includegraphics[width=0.22\textwidth]{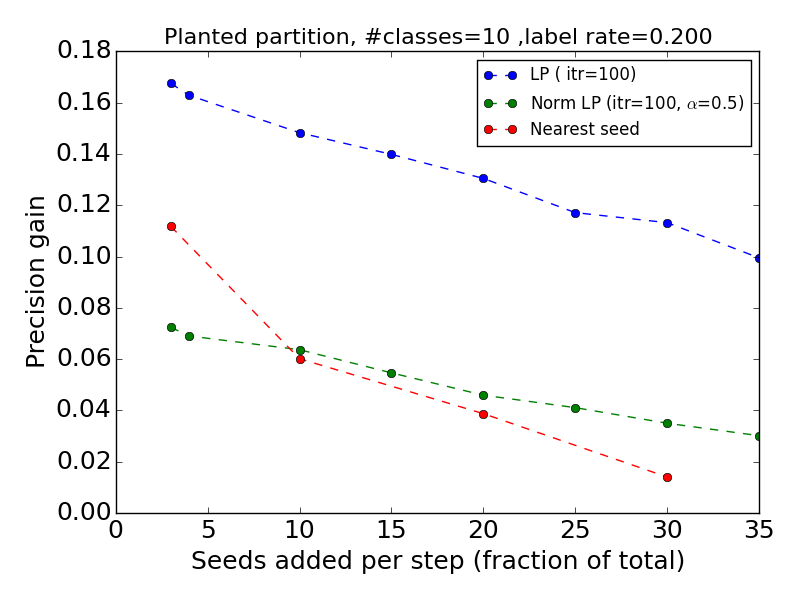}
}
\onlyinarxive{
  \includegraphics[width=0.22\textwidth]{cora_seeds_added_vs_precision_gain.png}
  \includegraphics[width=0.22\textwidth]{random_1_seeds_added_vs_precision_gain.png}
  \includegraphics[width=0.22\textwidth]{random_0_seeds_added_vs_precision_gain.png}
    \includegraphics[width=0.22\textwidth]{random_2_seeds_added_vs_precision_gain.png}
}
\caption{{\small Precision gain by bootstrapping as a function of the
    fraction of nodes (in percent) that
    are set as new seed nodes in each bootstrapping step}}
\label{precisionVSpercent:fig}
\end{figure}

We study the gain in precision as we sweep the bootstrapping
parameter $r$, which determines the fraction of nodes that are set as
seeds in each bootstrapping step (see Algorithm~\ref{bootstrap:alg}).
Figure~\ref{precisionVSpercent:fig} shows the precision gain as a
function of $r$ (in percent) for selected data sets. We can see that as expected the gain decreases with $r$  but that we can obtain significant gains also with relatively large values of $r$.

\subsection{Number of bootstrapping steps}
We study the precision as a function of the number of bootstrapping
steps performed.   
In our experiments we used the performance on a validation set to 
choose the step which provides the final learned labels.  Generally,
we expect precision to initially improve as the easier-to-predict nodes are added to the seed set and eventually to stabilize or decrease.

\begin{figure}
  \center
\notinarxive{
  \includegraphics[width=0.22\textwidth]{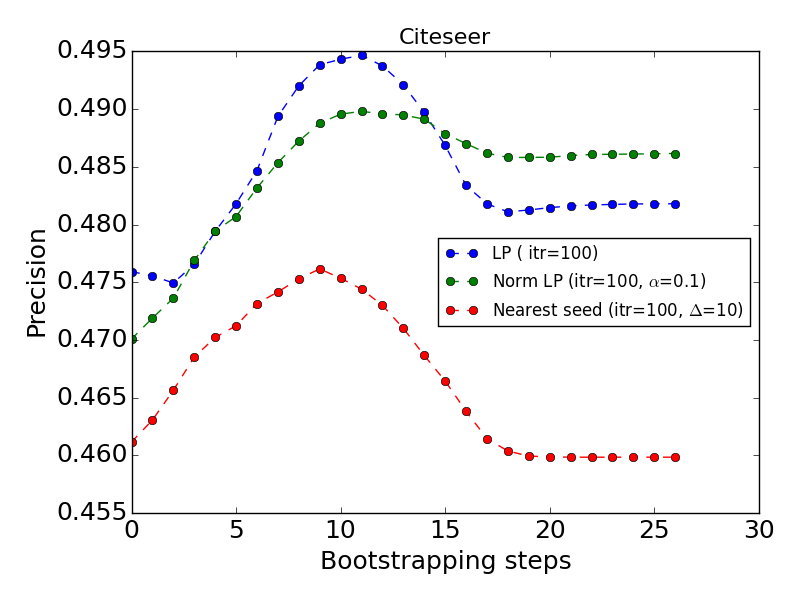}
  \includegraphics[width=0.22\textwidth]{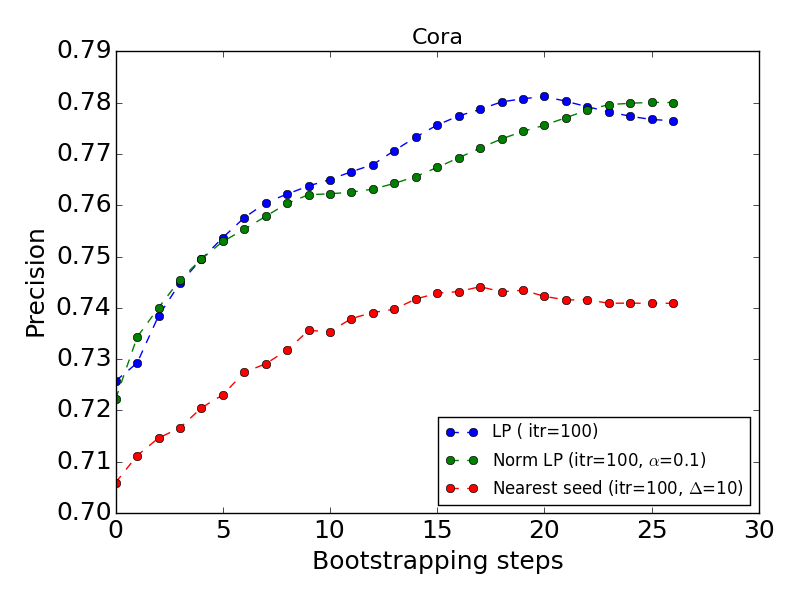}
  \includegraphics[width=0.22\textwidth]{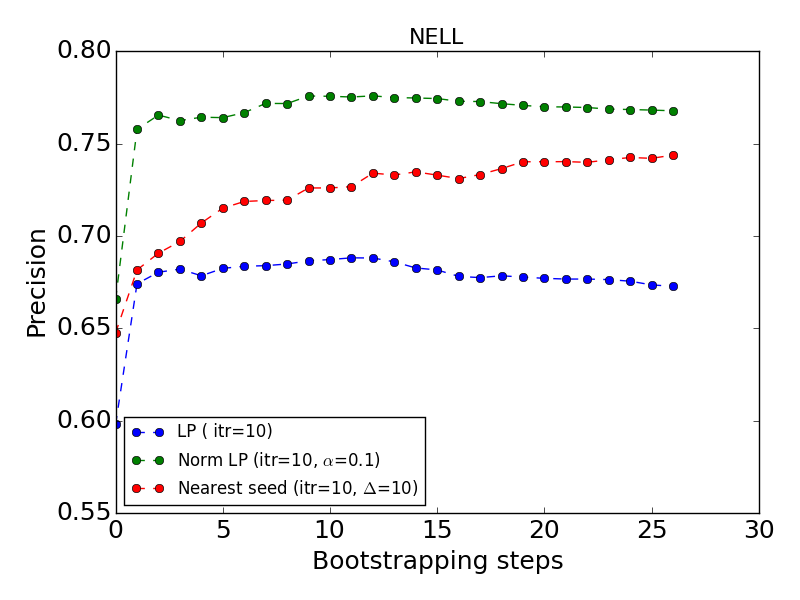}
    \includegraphics[width=0.22\textwidth]{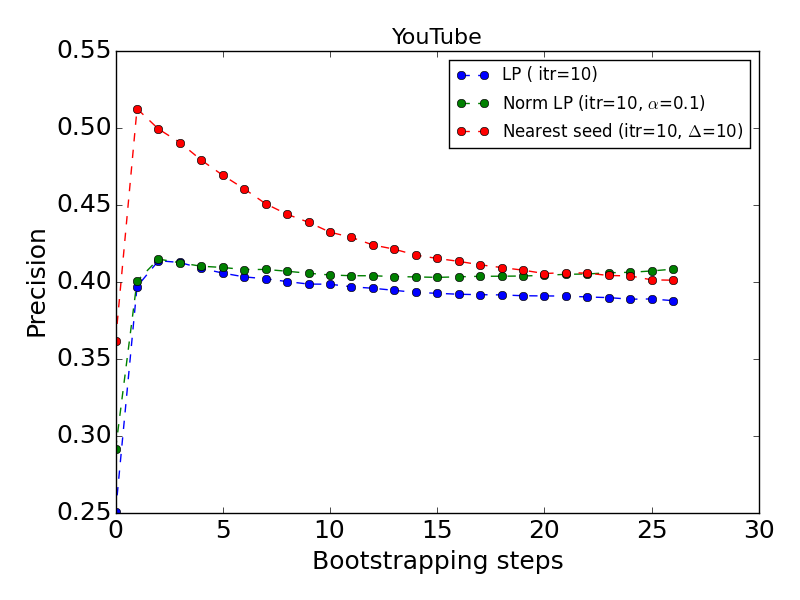}
  \includegraphics[width=0.22\textwidth]{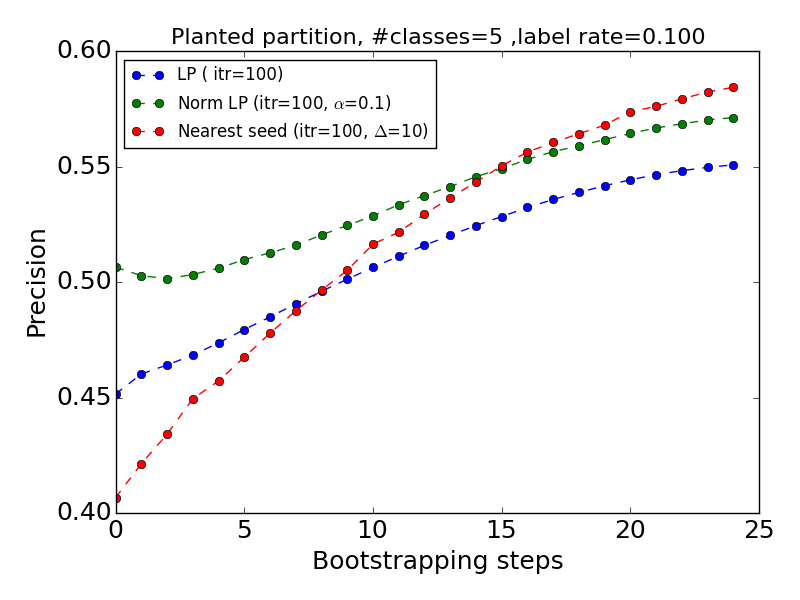}
    \includegraphics[width=0.22\textwidth]{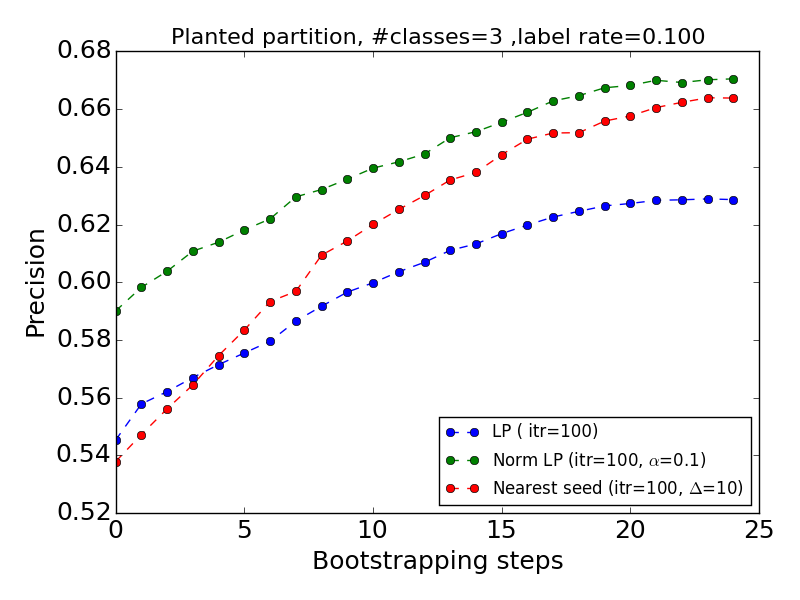}  
}
\onlyinarxive{
  \includegraphics[width=0.22\textwidth]{citeseer_bootstrapping_steps_vs_precision.png}
  \includegraphics[width=0.22\textwidth]{cora_bootstrapping_steps_vs_precision.png}
  \includegraphics[width=0.22\textwidth]{NELL_0_1_bootstrapping_steps_vs_precision.png}
    \includegraphics[width=0.22\textwidth]{youtube_bootstrapping_steps_vs_precision.png}
  \includegraphics[width=0.22\textwidth]{random_data_set_0_bootstrapping_steps_vs_precision.png}
    \includegraphics[width=0.22\textwidth]{random_data_set_1_bootstrapping_steps_vs_precision.png}  
}
\caption{{\small Precision as a function of the number of steps 
    performed by the bootstrapping wrapper.  We used $r=0.03$ and 
    fixed other (hyper)parameters as indicated in the legends.}}
\label{precisionVSsteps:fig}
\end{figure}

 Figure~\ref{precisionVSsteps:fig} shows the precision for each step  on representative data sets. In this set of experiments we fixed all other parameters of each
 algorithm as indicated in the legends of the figure and used a fixed
value $r=0.03$ for the fraction of nodes that become new seeds in each
step.  The results are averaged over 10 random splits of the data.
The Figure shows different  patterns but all are unimodal which means
they allow us to incorporate a stopping criteria for the bootstrapping algorithm.

%  Show precision/recall curves of sorted sequence, to explain the
%  bootstrapping.  Sort predictions by ``margin.'' 
%Show it for different steps.

%\begin{figure*}
%\center 
% \includegraphics[width=0.45\textwidth]{cora_graph_laplacian_seed_size_vs_accuracy.png}
%\caption{{\small Precision versus Label-rate for 
%    LP, Normalized Laplacian LP, and NN (regular and bootstrapped).   CORA.}}
%\label{precisionVSlabelingrateCORA:fig}
%\end{figure*}

\subsection{Scalability}

Optimizing quality is important when labeling is costly.  Often, however,
on very large graphs, scalability is critical.
The computation cost of  the methods we considered, social and spectral, bootstrapped or not,
depends on the total number of iterations performed.  

With label propagation, in each iteration for each node we compute an
average over its neighbors.   The averaging for different nodes in the
same iteration are independent and can be distributed or
parallelized.  But the iterations sequentially depend on each other.  The
computation of each iteration involves a linear number of edge traversals
and is highly suitable for Pregel-like \cite{pregel:sigmod2010}  distributed graph processing platforms.

With distance diffusion, each iteration is equivalent to a
single-source shortest-paths computation.  The number of edge
traversals performed is also linear.   
The iterations here are independent, and can be performed
concurrently, each providing independent samples from a distribution.
The shortest-path search performed in each iteration, however, has
concurrency that depends on the number of hops in the shortest-paths.
This computation can also be performed efficiently 
on Pregel \cite{pregel:sigmod2010} by essentially performing all
iterations (different sets of hash-specified edge lengths) together.
 With bootstrapping, the steps must be sequential and in each step
  we run the base algorithm with multiple iterations.  

%The total
%  computation depends on the total number of iterations across steps.

\begin{figure}
\center 
\notinarxive{
\includegraphics[width=0.22\textwidth]{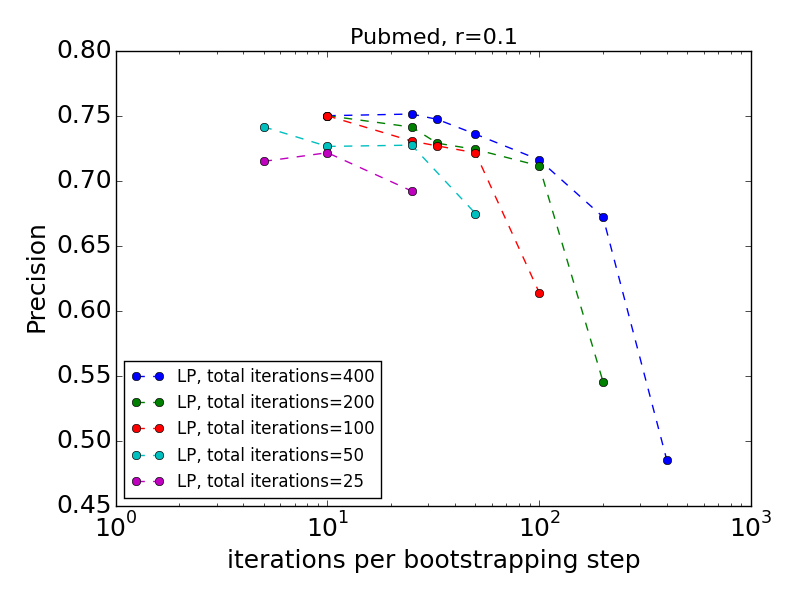}
\includegraphics[width=0.22\textwidth]{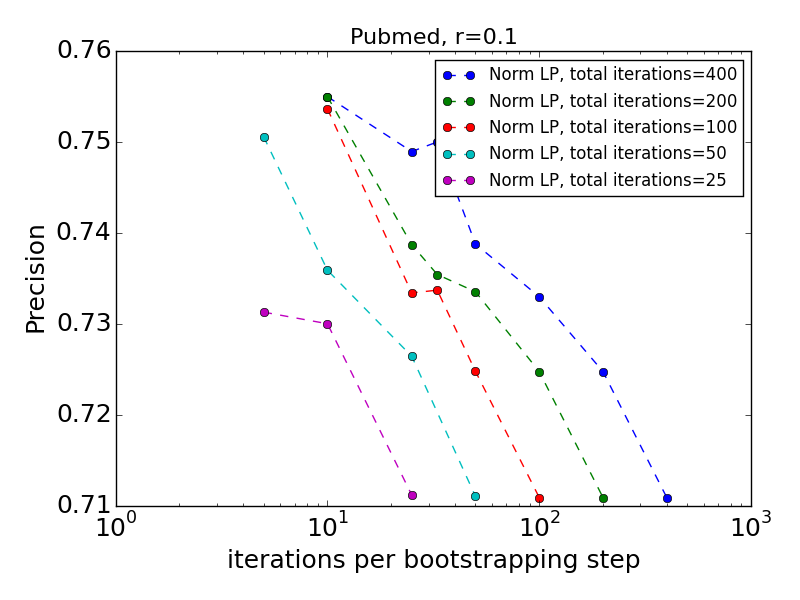}
\includegraphics[width=0.22\textwidth]{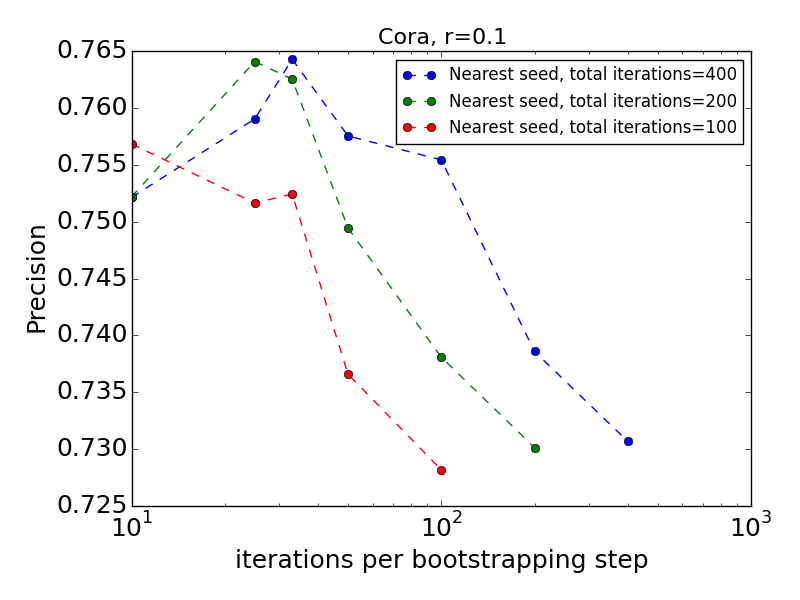}
\includegraphics[width=0.22\textwidth]{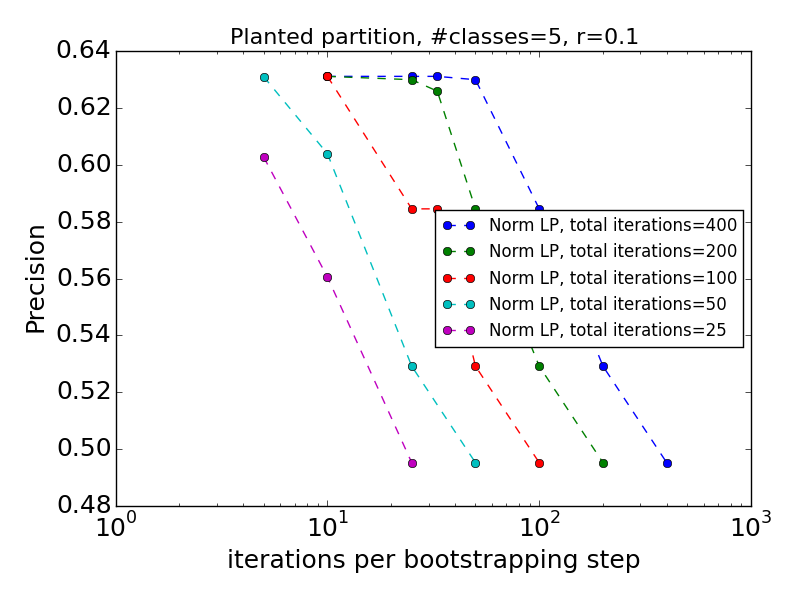}
\includegraphics[width=0.22\textwidth]{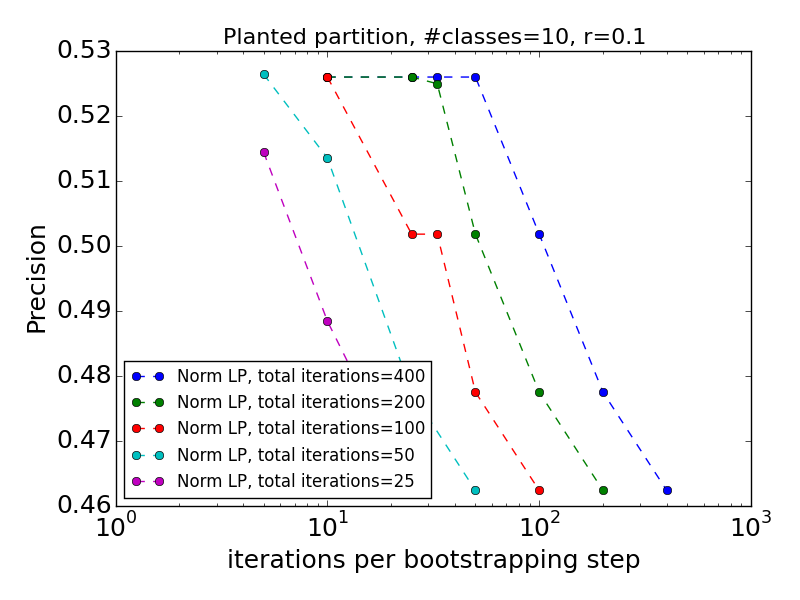}
\includegraphics[width=0.22\textwidth]{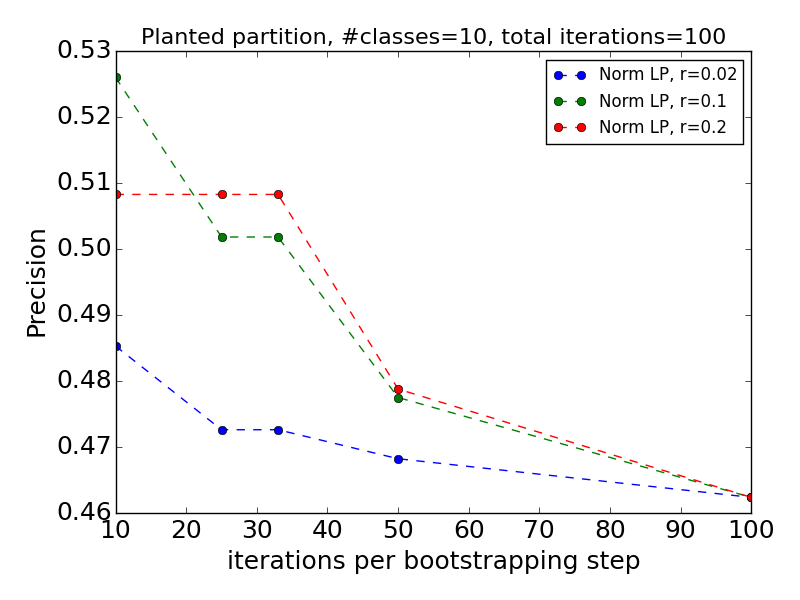}
}
\onlyinarxive{
\includegraphics[width=0.22\textwidth]{pubmed_laplacian_percent_10_vs_precision.png}
\includegraphics[width=0.22\textwidth]{pubmed_norm_laplacian_percent_10_vs_precision.png}
\includegraphics[width=0.22\textwidth]{cora_NN_percent_10_vs_precision.png}
\includegraphics[width=0.22\textwidth]{random_data_set_0_norm_laplacian_percent_10_vs_precision.png}
\includegraphics[width=0.22\textwidth]{random_data_set_2_norm_laplacian_percent_10_vs_precision.png}
\includegraphics[width=0.22\textwidth]{random_data_set_2_norm_laplacian_budget_100_vs_precision.png}
}
\caption{{\small Precision for a fixed iteration budget when varying
    the number of iterations per bootstrapping steps.}}
\label{fixedbudget:fig}
\end{figure}

 In this set of experiments we study the precision we obtain
 using a  fixed total number of iterations, for different sets of bootstrapping parameters.
  The plots in Figure~\ref{fixedbudget:fig} show precision as a function of
  the number of iterations performed in each  bootstrapping step.
  Each graph corresponds to a fixed budget of iterations (between 25
  and 400 iterations).   The number of iterations performed per step varies
  between $5$ and the full budget.
 When all iterations are performed in one step, that is, when the
 number of iterations per step is equal to the total budget,  we have
 the precision of the non-bootstrapped base algorithm.   
 When we partition the iterations budget 
  to multiple steps we reduce the effectiveness of the base algorithm 
  in each step but can leverage the power of bootstrapping.

Recall that the normalized Laplacian LP and nearest-seed improve with
more iterations per step whereas LP may not.  We can see that the
non-bootstrapped LP algorithm degrades with more iterations (recall that in other experiments we
treated the number of iterations with LP it as a hyperparameter).  The
plot for the Pubmed dataset show that with all budgets, quality picks
at 10 iterations per step.
With normalized Laplacian LP and nearest-seed (which only improve with
iterations), we see that
bootstrapped methods with 25 total iterations outperforms the 
non-bootstrapped algorithms with many more iterations. 

 Finally, we consider the parameter $r$, which determines the fraction
 of nodes that are instantiated as new seeds in each step.
  Generally bootstrapping performance improves with smaller values of
  the parameter $r$.  But with limited iteration budget, and limited
  number of steps,  very low values limit the progress for the bootstrapping.
The bottom right plot in Figure~\ref{fixedbudget:fig}  shows a sweet spot at
$r=0.1$.

\section{Feature diffusion}  \label{features:sec}
In the previous sections we focused on methods that learn labels using
only the graph structure:  The provided labels of seed nodes are
``diffused''  along graph edges to obtain soft labels for all nodes
that are then used for prediction.   When we have more information, in
the form of node feature vectors $\vecf_i$ for all
nodes $i\leq n_\ell+n_u$,  we can use it for
learning.  The simplest method, which does not use the graph
structure, is to train a supervised classifier
on the features and labels of seed nodes $(\tilde{\vecf}_i,y_i)$ $i\leq n_\ell$.
We can instead use the graph structure to obtain  {\em diffused
  feature vectors} $\tilde{\vecf}_i$.  The diffused vectors are
a smoothed version of the raw vectors that also reflect
features of related nodes, where relation is according to the
base diffusion process.
 Note that the diffused 
vectors $\tilde{\vecf}_i$ do not depend on the set of seed nodes or
their labels.
A supervised classifier $C$ can then be trained
on the diffused features and labels of seed nodes: $(\tilde{\vecf}_i,y_i)$ $i\leq n_\ell$.
When the classifier provides a prediction margin with each
classification $C(\tilde{\vecf_i})$ for $i>n_\ell$, it can be bootstrapped
(see algorithm \ref{bootstrap:alg}).

  We evaluated two diffusion methods.  {\em Feature Propagation} (FP), in 
  Algorithm \ref{fp:alg}, that uses the diffusion rule of the label propagation 
  method of \cite{ZhuGhahramani2002} (Algorithm \ref{lp:alg}), and {\em Normalized 
    Laplacian FP}, in Algorithm~\ref{nfp:alg}, that 
uses the diffusion rule of \cite{ZhouBLWS:NIPS2004} (Algorithm
\ref{ls:alg}).
% We use the notation $\vecf_{i}$ for the feature vector of node $i
% \leq n_\ell+n_u$.
We note that our linear feature diffusions can be viewed as a toned-down
GCN \cite{AtwoodT:NIPS2016,KipfW:ICLR2017}, without the backprop training, and non-linear
aggregations.

\begin{algorithm2e}[h]\caption{{\sc Feature Propagation (FP)} \label{fp:alg}}
  {\small
\SetKw{labels}{labels}
\DontPrintSemicolon
\KwIn{Affinity matrix $\boldW$. % Seed labels $y_i\in[L]$ for $i\in[n_\ell]$.
Node feature vectors $\vecf_i$ for $i\leq n_\ell+n_u$}
\tcp{Initialize:}
$\forall i,\ D_{ii}\gets \sum_j W_{ij}$\tcp*{diagonal degree matrix}
$\boldF \gets  (\vecf_1,\ldots,\vecf_{n_\ell+n_u})$\tcp*{Matrix of 
  feature vectors}
$\tilde{\boldF} \gets  \boldF$\tcp*{Diffused features matrix}
\tcp{Diffuse:}
\ForEach{$t=1,\ldots,T$}{
$\tilde{\boldF} \gets \boldD^{-1} \boldW \tilde{\boldF}$\;
}
\tcp{Finalize:}
\Return $(\tilde{\vecf}_1,\ldots,\tilde{\vecf}_{n_\ell+n_u}) \gets
\tilde{\boldF}$\tcp*{Diffused feature vectors}
% Train classifier $C$ on  labeled examples $(\tilde{\vecf}_i,y_i)$ $i\leq n_\ell$\;
% \Return classification $C(\tilde{\vecf}_i)$ and margin for $i> n_\ell$
}
\end{algorithm2e}

\begin{algorithm2e}[h]\caption{{\sc Normalized Laplacian FP} \label{nfp:alg}}
  {\small
\SetKw{labels}{labels}
\DontPrintSemicolon
\KwIn{Affinity matrix $\boldW$. % Seed labels $y_i\in[L]$ for $i\in[n_\ell]$.
Node feature vectors $\vecf_i$ for $i\leq n_\ell+n_u$. Parameter $\alpha\in (0,1)$}
\tcp{Initialize:}
$\forall i,\ D_{ii}\gets \sum_j W_{ij}$\tcp*{diagonal degree matrix}
$\boldF \gets  (\vecf_1,\ldots,\vecf_{n_\ell+n_u})$\tcp*{Matrix of 
  feature vectors}
$\tilde{\boldF} \gets  \boldF$\tcp*{Diffused features matrix}
$\boldA \gets \boldD^{-1/2} \boldW \boldD^{-1/2}$\tcp*{Normalized adjacency matrix}
\tcp{Diffuse:}
\ForEach{$t=0,\ldots,T$}{
$\tilde{\boldF}\gets (1-\alpha) \boldA \tilde{\boldF} + \alpha
\boldF$\;}
\Return $(\tilde{\vecf}_1,\ldots,\tilde{\vecf}_{n_\ell+n_u}) \gets
\tilde{\boldF}$\tcp*{Diffused feature vectors}
% Train classifier $C$ on  labeled examples $(\tilde{\vecf}_i,y_i)$ $i\leq n_\ell$\;
% \Return classification $C(\tilde{\vecf}_i)$ and margin for $i> n_\ell$
}
\end{algorithm2e}

\subsection{Experiments settings}
We  used the three citation networks dataset ({\sc Citeseer}, {\sc Cora}, and
{\sc Pubmed}) listed in Table~\ref{data:tab}.
We use the methodology of Section~\ref{experiments:sec} for the
selection of training, test, and validations sets.  
We use
the benchmark fixed seed sets used in prior work \cite{Yang:ICML2016,KipfW:ICLR2017},
to facilitate comparison, and random splits for robustness.  
The citation networks contain a bag-of-words
representation for each document which, following ~\cite{Yang:ICML2016,KipfW:ICLR2017}, we treat as a feature vector. The vectors are encoded using 0/1 which indicates the absence/presence of the corresponding term from the dictionary. The dictionaries of {\sc Citeseer}, {\sc Cora} and {\sc Pubmed} contain 3703, 1433 and 500 unique words respectively.
For classification from (original and diffused) feature vectors we used
one-versus-all logistic regression with the
Python sklearn library implementation \footnote{\url{http://scikit-learn.org/stable/about.html}}.

We  evaluate the classification quality when using the 
raw feature vectors $\vecf_i$ and when using diffused feature
vectors $\tilde{\vecf}_i$ obtained using Algorithms \ref{fp:alg} and Algorithm \ref{nfp:alg}.
For each base algorithm we also apply the
bootstrapping wrapper. 
% For comparison, we list the respective results reported using Graph
% Convolutional Networks (GCN)
% \cite{KipfW:ICLR2017} and Planetoid (best
% version) \cite{Yang:ICML2016}.  
We used a range of hyper/parameters with a validation set to prevent overfitting:
The bootstrapping wrapper was used with $r\in\{0.05,0.1,0.2\}$
as the fraction of new nodes selected as seeds in each step (see Algorithm~\ref{bootstrap:alg}).
For FP and normalized Laplacian FP  we used
$\{2, 5, 10, 20, 40, 100, 200\}$ propagation iterations to compute the diffused
feature vectors.  For normalized Laplacian FP we used $\alpha\in
\{0.01,0.05,0.1, 0.2\}$.  We comment that the best results across data
sets were obtained with $10$ iterations and with $\alpha=0.2$.

\subsection{Results on benchmark datasets}
Results on benchmark datasets are reported in
% In this set of experiments we compare the algorithms when running on
% a fixed seed set used in prior  work \cite{Yang:ICML2016}. Results
% are displayed in table 
Table~\ref{feature_bootstrapped_benchmark:tab}.  The table also lists for
reference the results obtained without using node features by
bootstrapped normalized Laplacian LP (experiments in
Section~\ref{benchmark:sec}).   For comparison, we also list the
quality reported by {\sc GCN} \cite{KipfW:ICLR2017}  and {\sc
  Planetoid} \cite{Yang:ICML2016} (best variant with node features).

We observe the following:
First,  the quality of learning with diffused feature vectors is significantly
better than with the raw feature vectors, with average
improvement of about $12\%$.   Hence in these data sets the use of the
graph structure and the particular way it was used were important.
Second, the normalized Laplacian FP was more
effective than basic FP.  This agrees with our
observations with the label propagation experiments.
Third,
bootstrapping consistently improved performance on two of the data
sets ({\sc Citeseer} and {\sc Cora}).  There was little or no
improvement on {\sc Pubmed}, but on that data sets bootstrapped label
propagation (that did not used the node features) was the near-best
performer.
Fourth, the bootstrapped version of normalized Laplacian FP
improves over the state of the art results of {\sc GCN} \cite{KipfW:ICLR2017} on
{\sc Citeseer} and {\sc Cora}.  On {\sc Pubmed} {\sc GCN} was only slightly
better than our bootstrapped label propagation.

% We can also see that the normalized versions consistently outperform
% the non normalized versions in label and feature propagation. On
% {\sc Pubmed} GCN \cite{KipfW:ICLR20% 17} and normalized label propagation [\ref{ls:alg}] obtain the best results with a small advantage to GCN. In most cases bootstrapping increased the accuracy of base algorithm.

\subsection{Results on random splits}
In this set of experiments, for each data set, we generated multiple random splits of 
 the nodes to seed, test, and validation sets and averaged the results.
Our results are reported in Table~\ref{feature_bootstrapped:tab}. 
For reference, we also list the results using label
propagation (without the use of node features) that we reported in
Section~\ref{randomsplits:sec} and the results reported on similar
random splits using GCNs \cite{KipfW:ICLR2017}.
The results add robustness to our observations from the benchmark
experiments:  The use of diffused features significantly improves
quality, the normalized Laplacian FP consistently achieves the best
results on {\sc Citeseer} and {\sc Cora} and is very close (within
error margins) to the results reported by \cite{KipfW:ICLR2017} on {\sc Pubmed}.

\begin{table}[!ht]  \caption{Feature diffusion results for benchmark data \label{feature_bootstrapped_benchmark:tab}}
\center 
{\small
\begin{tabular}{c|lll}
Method & {\sc Citeseer} & {\sc Cora} & {\sc Pubmed} \\
\hline\hline
{\sc Norm Lap LP}\cite{ZhouBLWS:NIPS2004} {\sc +Bootstrapped}  & 0.536 & 0.784  & 0.788 \\
\hline
\hline
{\sc no feature diffusion} & 0.604 & 0.589 & 0.729 \\ 
 \textcolor{blue}{+Bootstrapped}  & 0.683 & 0.655 & 0.729 \\ 
 \hline 

{\sc Feature propagation} & 0.687 & 0.804 & 0.779 \\ 
 \textcolor{blue}{+Bootstrapped}  & 0.703 & 0.798 & 0.711 \\ 
 \hline 

{\sc Norm feature propagation} & 0.696 & 0.824 & 0.765 \\ 
 \textcolor{blue}{+Bootstrapped}  & {\bf 0.728} & {\bf  0.829} & 0.765 \\ 
  
\hline 
{\sc Graph Conv Nets} \cite{KipfW:ICLR2017}  & 0.703 & 0.815 & {\bf 0.790} \\ 
 \hline
{\sc Planetoid} \cite{Yang:ICML2016} & 0.629 & 0.757 & 0.757 \\ 
 \hline
\end{tabular}
}
\end{table}

\begin{table}[!ht]  \caption{Feature diffusion results over random splits \label{feature_bootstrapped:tab}}
\center 
{\small
\begin{tabular}{c|lll}
Method & {\sc Citeseer} & {\sc Cora} & {\sc Pubmed} \\
\hline\hline
$\times$ repetitions   & 100  & 100  & 100    \\
\hline

{\sc Label Propagation}\cite{ZhuGhahramani2002} & 0.479  & 0.728  & 0.709 \\
\textcolor{blue}{+Bootstrapped} & 0.496 & 0.781 & 0.747 \\
\hline
{\sc Norm Lap LP} \cite{ZhouBLWS:NIPS2004}  & 0.490 & 0.730  & 0.739 \\
\textcolor{blue}{+Bootstrapped} & 0.503 & 0.782 &  0.756 \\
\hline
\hline
$\times$ repetitions   & 10  & 10  & 10    \\
\hline

{\sc no feature diffusion} & 0.614 & 0.608 & 0.719 \\ 
 \textcolor{blue}{+Bootstrapped}  & 0.695 & 0.695 & 0.757 \\ 
 \hline 

{\sc Feature propagation} & 0.703 & 0.808 & 0.781 \\ 
 \textcolor{blue}{+Bootstrapped}  & 0.719 & 0.832 & 0.781 \\ 
 \hline 

{\sc Norm feature propagation} & 0.727 & 0.831 & 0.785 \\ 
 \textcolor{blue}{+Bootstrapped}  & {\bf 0.737} & {\bf 0.848} &  0.785 \\ 
 \hline 
 \hline 
{\sc Graph Conv Nets} \cite{KipfW:ICLR2017}  & 0.679 & 0.801 & {\bf 0.789} \\ 
\end{tabular}
}
\end{table}

\section{Conclusion} \label{conclu:sec}

 We studied the application of self-training, which is perhaps the most basic form of introducing  non-linearity, to SSL methods based on linear graph diffusions. 
 We observed that the resulting {\em bootstrapped diffusions} ubiquitously
 improved labeling
quality over the respective base methods on a variety of real-world and
synthetic data sets.  Moreover, 
we obtain  state-of-the-art quality, previously achieved by more
complex methods, while retaining the high scalability of the base methods.

Our results are a proof of concept that uses the simplest base algorithms and bootstrapping wrapper.  Some natural extensions include
fine tuning of the wrapper together with base algorithms that provide more precise confidence scores and the use of a richer set of base algorithms.

On a final note, we recall that spectral and social graph
diffusions are an important tool in graph mining:  They are
the basis of centrality, influence, and
similarity  measures of a node or sets of nodes in a network
\cite{Chung:Book97a,pagerank:1999,KKT:KDD2003,BlochJackson:2007,CoKa:jcss07,Opsahl:2010,GRLK:KDD2010,CDFGGW:COSN2013,DSGZ:nips2013,timedinfluence:2015}
and also underline community detection and local clustering
algorithms~\cite{SpielmanTeng:sicomp2013}.  Our work suggests that
substantial gains in quality might be possible by using
bootstrapped diffusions as an
alternative to classic ones in these wider contexts. We hope to pursue
this in future work.

%Finally, we note that with the abundance of complex new learning
%methods, our work is an example where one simple ingredient can be credited with much of the 
%  demonstrated gain in quality.   

%  Place back other components: node features.
%  We only used the most basic form of bootstrapping.  Look into 
% Incorporate in the diffusions other components that undeline the success of newer learning
%  algorithm.  Dropout.  

\begin{acks}
  The authors would like to thank 
Aditya Grover,  the author 
  of {\sc node2vec} \cite{node2vec:kdd2016} and
Zhilin Yang, the author of
  {\sc Planetoid}~\cite{Yang:ICML2016} 
for
  prompt and helpful answers to our questions on their work and
  implementations, and to Fernando Pereira for his advice.
%   The first author is 
This research is partially
  supported by the \grantsponsor{ccc}{Israel Science Foundation}\/ (Grant
  No.~\grantnum{}{1841/14}).
\end{acks}

\bibliographystyle{plain}
\bibliography{cycle}

\end{document}